\documentclass[11pt,a4paper]{article}
\usepackage{pasteurlabs}

\usepackage{nicefrac}
\usepackage[normalem]{ulem}
\usepackage{colortbl}

\definecolor{firstbg}{rgb}{0.68, 0.82, 0.96}
\definecolor{secondbg}{rgb}{0.85, 0.91, 0.97}

\newcommand{\sn}[2]{#1\,{$\cdot$}\,10\textsuperscript{--#2}}
\newcommand{\first}[1]{\cellcolor{firstbg}#1}
\newcommand{\second}[1]{\cellcolor{secondbg}#1}

\newcommand{\R}{\mathbb{R}}
\newcommand{\Xset}{\mathcal{X}}
\newcommand{\Pset}{\mathcal{P}}

\newcommand{\Enc}{E}

\newcommand{\SelfAttn}{\mathrm{SelfAttn}}
\newcommand{\CrossAttn}{\mathrm{CrossAttn}}

\newcommand{\softmax}{\mathrm{softmax}}
\newcommand{\LayerNorm}{\mathrm{LayerNorm}}
\newcommand{\MLP}{\mathrm{MLP}}

\title{Courant: a State-Adaptive Perceiver-Based Neural Surrogate with Local Support and Interpretable Field Decomposition}

\author{Anuj Kumar\textsuperscript{1,*},\;
        Josiah Bjorgaard\textsuperscript{1,*},\;
        Nikolaos Bouklas\textsuperscript{1,2},\;
        Matteo Salvador\textsuperscript{1},\;
        Alexander Lavin\textsuperscript{1,3}}

\date{May 2026}

\affiliation{%
  \textsuperscript{1}Pasteur Labs, NY, NY, USA\qquad
  \textsuperscript{2}Cornell University, Ithaca, NY, USA\\[2pt]
  \textsuperscript{3}Institute for Simulation Intelligence, NY, NY, USA\\[2pt]
  $^{*}$Equal contribution.}

\correspondence{lavin@simulation.science}

\organization{\includegraphics[height=26pt]{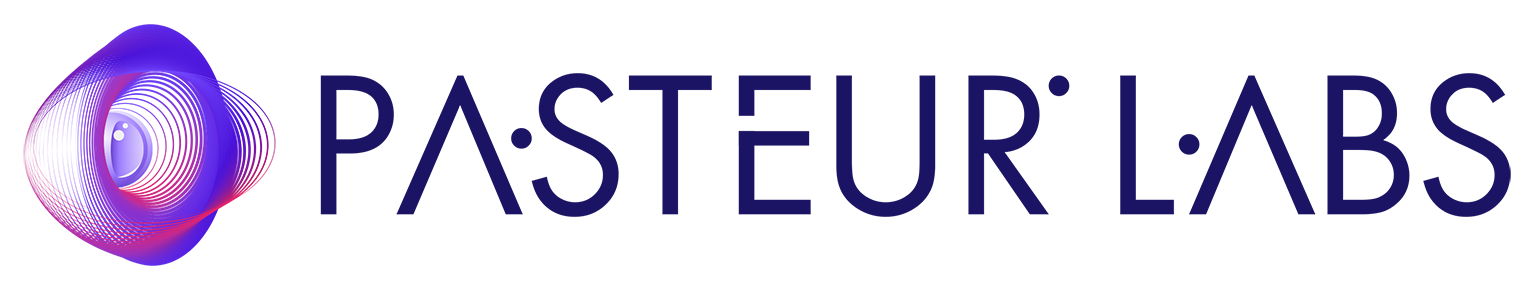}}

\shorttitle{Courant: a State-Adaptive Perceiver-Based Neural Surrogate with Local Support}

\begin{document}

\maketitle

\begin{plabstract}
We introduce ``Courant'', a Perceiver-based encoder--processor--decoder
surrogate model that has latent features exhibiting adaptive
specialization and local support in the physical space, enabling
functionality akin to an adaptive \textit{hp}-refinement scheme—an
attribute that is highly desirable in traditional numerical solvers
and scientific machine learning broadly. 
The proposed architecture combines a
shared random Fourier feature coordinate embedding, state-adapted
latent queries, and a light-weight decoder. Courant is trained
end-to-end with steady or transient simulation data and only a
standard L$_2$ prediction loss in the physical space, achieving
competitive accuracy on benchmarks. We demonstrate that Courant's
inductive biases yield latents that are interpretable by design: they
develop multiscale geometric specialization in the simulation domain
and track coherent structures in the time-dependent case, acting
analogously to time-evolving spatial basis functions and allowing for
decoding a compact, geometry-anchored, partition-of-unity-like
decomposition of the simulated field.
\end{plabstract}

\section{Introduction}\label{sec:introduction}

Numerical simulation of Partial Differential Equations (PDEs) is
fundamental to engineering physics, scientific discovery, and industrial systems and risk assessment, yet remains computationally prohibitive for design/solution-space exploration, uncertainty quantification, cause-effect reasoning, and real-time decision-making. Neural surrogates, trained to approximate the input-output map of a simulator, offer orders-of-magnitude speedups by amortizing the cost of repeated solves. Recent architectures based
on Graph Neural Networks
(GNNs)~\citep{pfaff2021learning,brandstetter2022message}, neural
operators~\citep{li2021fourier,lu2021learning}, and transformer-based
models~\citep{alkin2024universal,hagnberger2025calm,wu2024transolver}
have demonstrated competitive accuracy on increasingly complex
benchmarks, although not without inconsistencies and over-simplifications~\citep{McGreivy2024WeakBA}. Motivated by the recent rise of physics-informed machine
learning, there is growing interest in hybridizing neural
architectures with concepts from traditional numerical solvers and in
enhancing the interpretability of neural surrogates, which are
commonly viewed as black
boxes~\citep{karniadakis2021physics,toscano2025pinns,mandl2025separable}.

In applied physics and engineering simulation, professionals rely on physical intuition to validate results, diagnose failures, and build trust in
computational tools. A surrogate that produces accurate predictions
through an inscrutable internal process offers weaker guarantees than
one whose internal representations can be inspected and related to
known physical structure. Interpretability in this context does not
require full mechanistic transparency; it requires that a surrogate
model's internal organization be legible enough to support diagnosis,
validation, comparison with established computational methods like the
Finite Element Method (FEM) and domain decomposition, and enable
hybridization towards bridging the gap from solvers to surrogates.
These features could be used to offer new modes of human-machine
teaming and partially define a surrogate model's technology readiness
level in real-world use-cases~\citep{lavin2022trl,lavin2021simulation}.

A crucial step in the automation of the numerical solution of PDEs in
complex domains was the departure from the Rayleigh-Ritz method, where
the construction of global basis functions was necessary, to the
introduction of the FEM~\citep{liu2022eighty} where local support was
inherently embedded in the construction of basis
functions~\citep{courant1943variational}. Mesh and polynomial order
adaptivity, often referred to as \textit{hp}-adaptivity, was also
crucial in solving complex problems while maintaining computational
cost~\citep{guo1986hp}. In neural surrogates, this notion of local
support and spatial adaptivity is lost, as these approaches commonly
construct global approximations that are not easily decomposed, beyond
spatial decomposition in subdomains~\citep{jagtap2020extended} and
some other cases~\citep{lee2021partition,tripura2022wavelet}.

Classical reduced-order modeling also provides a useful reference
point. Proper Orthogonal Decomposition (POD) constructs a low-rank
spatial basis from simulation snapshots and represents the solution as
a linear combination of these modes, each with a time-varying
coefficient~\citep{berkooz1993proper}. Dynamic Mode Decomposition
(DMD) identifies spatiotemporal modes with associated frequencies and
growth rates~\citep{schmid2010dynamic}. These methods are
interpretable and decomposable by construction: the spatial modes can
be visualized, the temporal coefficients tracked, and the
representation directly compared with the full-field solution.
However, these methods operate on a linear subspace that is
effectively tangent to the underlying solution manifold, so their
representational capacity is bounded by the rank of the snapshot
matrix, limiting their effectiveness for strongly nonlinear or
advection-dominated dynamics.
Neural surrogates face no such linearity constraint, but their learned
representations are not structured to be interpretable by default let alone reliably sampled.

We seek surrogacy architectures providing multiphysics and multiscale models guided by known inductive biases associated with locality, sparsity, decomposability and adaptivity, with the goal of enabling adaptive, state-anchored latents with local support. 
In particular, random Fourier embeddings are known to encode coherent multiscale structure, geometry-anchored queries and cross-attention naturally promote localized feature extraction, Perceiver latent bottlenecks encourage sparse and interpretable latent organization, self-attention processors enable global communication between localized latent features, and affine decoders preserve the learned latent structure without corrupting feature specialization through additional nonlinear transformations. 

\textbf{To this end, we designed Courant, a
Perceiver-based~\citep{jaegle2021perceiver}
encoder--processor--decoder surrogate model whose design is jointly
motivated by predictive accuracy and latent legibility}:
geometry-anchored cross-attention, a Neural Ordinary Differential
Equation (NeuralODE) processor~\citep{chen2018neuralode}, and a
lightweight linearly-decomposing decoder sharing its Random Fourier
Feature (RFF) coordinate embedding with the encoder. Trained
end-to-end with only a standard $L_2$ prediction loss, Courant
develops latent structure reminiscent of partition-of-unity
decompositions: spatially localized, state adaptive, and temporally
coherent. Our primary contributions are:
\begin{enumerate}
    \item \textbf{An architecture for legible surrogates.} An
          architecture with design choices that are jointly motivated
          by predictive accuracy and latent legibility: state-adaptive
          cross-attention, shared random Fourier coordinate embedding
          anchoring encoder and decoder to a common geometric frame, a
          NeuralODE processor using self-attention for continuous latent dynamics; and a
          decoupled input--output design with design-parameter
          conditioning at the encoder and distance-field aware queries
          at the decoder.
    \item \textbf{Per-token field decomposition.} The decoder is
          deliberately kept affine in the attention output, which
          guarantees that the predicted field decomposes exactly into
          per-token contributions, each interpretable as a spatially
          localized, geometry-grounded basis function.
    \item \textbf{Emergent local support without structural
          supervision.} Trained end-to-end with only a standard
          data-driven prediction loss in the physical space---no
          sparsity, locality, or structure-inducing terms---Courant's
          latent tokens self-organize into an adaptive basis whose
          contributions tile the domain in a partition-of-unity-like
          decomposition. Under NeuralODE rollout, these tokens track
          coherent physical structures across time.
\end{enumerate}

\section{Related Work}

\paragraph{Neural surrogate models}
GNN surrogates~\citep{pfaff2021learning,brandstetter2022message}
operate on mesh-based representations and achieve strong results on
unsteady fluid dynamics, while neural operators like Fourier Neural
Operator (FNO)~\citep{li2021fourier} and
DeepONet~\citep{lu2021learning} learn mappings between function spaces
directly. Transformer-based approaches have seen widespread use. The
Transolver family~\citep{wu2024transolver,luo2025transolver++,
zhou2026transolver} compresses mesh points into latent tokens via
data-dependent soft assignment to learnable slices, applying
self-attention over those tokens. The Universal Physics Transformer
(UPT)~\citep{alkin2024universal} adapts the Perceiver pattern to
physics simulation, compressing point-cloud inputs into fixed-size
latent tokens via a hierarchical encoder and decoding via
cross-attention queries; related Perceiver-style
surrogates~\citep{alkin2024universal,serrano2024aroma} similarly
decouple input encoding from output query evaluation, though they
typically process the full input mesh. Geometry-aware operator
transformers~\citep{wen2025geometry} combine cross-attention encoding
with transformer processing on arbitrary domains, and Equivariant
Neural Fields (ENF)~\citep{wessels2025enf} use distance-based
bi-invariant cross-attention to associate each latent feature with a
spatial position. Latent Dynamics Networks~\citep{regazzoni2024learning}
jointly discover low-dimensional manifolds and dynamics for
spatio-temporal processes.

\paragraph{Interpretability in neural surrogate models}
As sparse discovery usually enables interpretability in the context of
data-driven learning~\citep{brunton2016discovering}, neural surrogate
models do not commonly accommodate such tasks.
\citet{barwey2025interpretable} add differentiable Top-K pooling to
GNN surrogates, producing spatial masks that serve as a-posteriori
error indicators.
FIGNN~\citep{raut2025fignn} extends this to feature-specific masks.
\citet{kim2025towards} use Discrete Empirical Interpolation Method (DEIM) trajectories to diagnose NeuralODE
dynamics, finding that interpolation points track physically meaningful
flow structures. Additionally, \citet{hu2025interpreting} utilize
sparse autoencoders to extract a dictionary of interpretable latent
features.

\paragraph{Representation decomposition in neural surrogate models}
The main form of interpreting predictions in neural surrogate models
is the decomposition of solution fields into a reduced basis on which
the core computation operates. FNO~\citep{li2021fourier} parameterizes
the integral kernel in a fixed Fourier basis, truncated to a set
number of modes. DeepONet~\citep{lu2021learning} implicitly learns a
nonlinear spatial basis through its trunk network, with
POD-DeepONet~\citep{lu2022comprehensive} making this connection
explicit by replacing the trunk with POD modes from training data.
Latent spectral model~\citep{wu2023solving} projects inputs into a
compact latent space and decomposes the operator into multiple learned
basis operators inspired by classical spectral methods.
GeoTransolver~\citep{adams2025geotransolver} recognizes that these
globally-supported slices lack geometric grounding and augments
Transolver's physics-attention with persistent cross-attention to
multi-scale ball-query geometry features at every layer, injecting
spatial structure as a conditioning signal while retaining the global
slice decomposition as the primary basis. More broadly in physics
informed machine learning, domain decomposition approaches have also
incorporated local support in representation, but commonly do not
offer adaptivity as seen in \citet{howard2024finite}.

\section{Model}\label{sec:model}

\begin{figure}[htb]
    \centering
    \includegraphics[width=\linewidth]{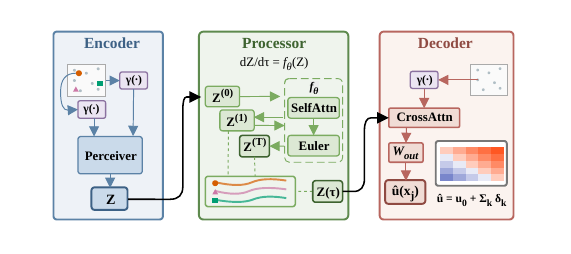}
    \caption{Design of Courant across encoder, processor, and decoder.
    Selected point anchors are used to compress the input point cloud
    into latents using Perceiver layers. The processor evolves the
    latent anchors through a NeuralODE using self-attention for the
    right hand side operator. The decoder is a separate cross attention
    layer followed by a linear layer. At inference time, the latents
    can be decoded separately.}
    \label{fig:architecture}
\end{figure}

As detailed in Figure \ref{fig:architecture}, our architecture follows
the encoder--processor--decoder
paradigm~\citep{alkin2024universal,serrano2024aroma,wang2024latent,
bekar2025hybrid,hagnberger2025calm}. Let the input point cloud at time
$t$ be defined by
$\Xset^t\!=\!\{(\mathbf{x}_i^t,\mathbf{f}_i^t)\}_{i=1}^{N}$ with
$\mathbf{x}_i\!\in\!\R^{d_c}$ and features $\mathbf{f}_i\!\in\!\R^{d_f}$
at every location. In practice, features include global (design)
parameters and a distance field to boundaries, where available. Let us
also define a set of $L\!\ll\!N$ anchor positions
$\Pset\!=\!\{\mathbf{p}_j\}_{j=1}^L$; the encoder produces a latent
matrix $\mathbf{Z}\!=\!\Enc(\Pset;\Xset)\!\in\!\R^{L\times d}$; the
decoder acts on $\mathbf{Z}$ directly (steady-state) or on the latent
trajectory $\mathbf{Z}(\tau)$ produced by a processor $\Phi$ where
$\tau$ is a set of $T-1$ discrete timesteps
$\{t+\Delta t, t+2\Delta t,\ldots, t+T\Delta t\}$.

\subsection{Shared random Fourier feature embedding}
\label{sec:rff}

Coordinates enter Courant exclusively through a shared RFF
map~\citep{tancik2020fourier,mildenhall2020nerf}
\begin{equation}
\gamma(\mathbf{x}) = \bigl[\cos(2\pi\,\sigma\,\mathbf{B}\mathbf{x}),\;
                            \sin(2\pi\,\sigma\,\mathbf{B}\mathbf{x})\bigr]
                     \!\in\!\R^{d},
\quad \mathbf{B}\!\in\!\R^{(d/2)\times d_c},\;
      \mathbf{B}_{rk}\!\sim\!\mathcal{N}(0,1),
\label{eq:rff}
\end{equation}
where $\mathbf{B}$ is a learnable parameter matrix. A single $\gamma$
is instantiated once and used by every stage: to lift query positions
and input coordinates inside the encoder, and to lift query
coordinates inside the decoder.

\subsection{Encoder: multi-level Perceiver with anchored queries}
\label{sec:encoder}

The encoder is a Perceiver stack of $L_{\text{enc}}$ levels with
separate weights. Unlike the original Perceiver, whose latent queries
are abstract learned vectors, every latent vector is tied to an
explicit spatial anchor $\mathbf{p}_j\!\in\!\R^{d_c}$ through the RFF
embedding of that anchor:
\begin{equation}
\mathbf{q}_j^{(0)} = \gamma\!\bigl(\mathbf{p}_j^{(0)}\bigr)\in\R^d.
\label{eq:anchor-init}
\end{equation}

Key/value tokens concatenate embedded coordinates, raw features, and
an optional global conditioning vector $\mathbf{g}\!\in\!\R^d$
(e.g.\ design parameters):
\begin{equation}
\mathbf{c}_i = \bigl[\gamma(\mathbf{x}_i)+\mathbf{g},\;\mathbf{f}_i\bigr]
               \in\R^{d+d_f}.
\label{eq:kv-tokens}
\end{equation}
Each level $\ell$ then applies cross-attention against $\{\mathbf{c}_i\}$
followed by self-attention:
\begin{align}
\tilde{\mathbf{Z}}^{(\ell)} &= \mathbf{Z}^{(\ell-1)}
  + \CrossAttn^{(\ell)}\!\bigl(\mathbf{Z}^{(\ell-1)},\,\{\mathbf{c}_i\}\bigr),
\label{eq:enc-ca}\\
\mathbf{Z}^{(\ell)}   &= \tilde{\mathbf{Z}}^{(\ell)}
  + \SelfAttn^{(\ell)}\!\bigl(\tilde{\mathbf{Z}}^{(\ell)}\bigr),
\label{eq:enc-saff}
\end{align}
with $\mathbf{Z}^{(-1)}\!=\!\{\mathbf{q}_j^{(0)}\}$ and encoder output
$\mathbf{Z}\!=\!\mathbf{Z}^{(L_{\text{enc}}-1)}$. Here, we imply the
typical residual, layernorm, and feed-forward network in every cross
attention and self attention layer.

\subsection{Processor: self-attention NeuralODE}
\label{sec:processor}

For time-dependent problems, latent anchors are advanced in latent
space by a NeuralODE~\citep{chen2018neuralode}, as in
DINo~\citep{yin2023dino} and CORAL~\citep{yin2023coral}. Here, the
right-hand side is a self-attention block over latent anchors, rather
than a per-token $\MLP$:
\begin{equation}
\frac{d\mathbf{Z}}{d\tau}
= f_\theta\!\bigl(\mathbf{Z}(\tau)\bigr)
= \SelfAttn\!\bigl(\mathbf{Z}(\tau)\bigr),
\qquad
\mathbf{Z}(\tau{+}h)=\mathbf{Z}(\tau)+h\,f_\theta\!\bigl(\mathbf{Z}(\tau)\bigr).
\label{eq:ode}
\end{equation}
The solution is rolled-out autoregressively with explicit Euler
integration over $T$ steps from
$\mathbf{Z}(0)\!=\!\Enc(\Pset,\Xset^{t_0})$.

\subsection{Decoder: minimal neural field with shared embedding}
\label{sec:decoder}

The decoder is effectively a neural
field~\citep{mildenhall2020nerf,yin2023coral,chen2023crom} conditioned
on $\mathbf{Z}$ via a single multi-head cross-attention with no
feed-forward network. Given a query coordinate $\mathbf{x}_j$ and
optional extra features $\boldsymbol{\xi}_j$ (e.g.\ distance-field),
the decoder embeds the query through the same $\gamma$ as the encoder
with additional augmentation,
$\mathbf{e}_j\!=\!\MLP\!\bigl([\gamma(\mathbf{x}_j),\,
W_\xi\boldsymbol{\xi}_j]\bigr)\!\in\!\R^d$,
and computes per-head attention weights
\begin{equation}
w_{h,k}(\mathbf{x}_j)
= \softmax_k\!\!\left(
  \frac{(W_Q^h\,\mathbf{e}_j)^{\!\top}(W_K^h\,\tilde{\mathbf{z}}_k)}
       {\sqrt{d_h}}
\right),
\qquad h=1,\dots,H,
\label{eq:dec-weights}
\end{equation}
where $\tilde{\mathbf{z}}_k\!=\!\LayerNorm(\mathbf{z}_k)$,
$W_Q^h,W_K^h,W_V^h\!\in\!\R^{d_h\times d}$ are the per-head query,
key, and value projections. The predicted field is then
\begin{equation}
\hat{\mathbf{u}}(\mathbf{x}_j)
= W_{\mathrm{out}}\!\sum_{h=1}^{H} W_O^h
  \sum_{k=1}^{L} w_{h,k}(\mathbf{x}_j)\;W_V^h\,\tilde{\mathbf{z}}_k,
\label{eq:dec-output}
\end{equation}
where $W_O^h\!\in\!\R^{d\times d_h}$ is the $h$-th row block of the
multi-head output projection $W_O\!\in\!\R^{d\times Hd_h}$, and
$W_{\mathrm{out}}\!\in\!\R^{d_{\mathrm{out}}\times d}$ maps to the
physical output dimension. This design is deliberately minimal so that
it evaluates cheaply at millions of query points and relies on the
representational capacity of the encoder.

\subsection{Partition of unity and per-anchor decomposition}

For each head $h$, the weights satisfy $w_{h,k}\!\geq\!0$ and
$\sum_{k}w_{h,k}(\mathbf{x}_j)\!=\!1$ with compact spatial support,
forming a partition of unity over the anchor cloud. Because the
decoder is affine in the value vectors, the predicted field decomposes
exactly into spatially localized per-anchor contributions:
\begin{equation}
\hat{\mathbf{u}}(\mathbf{x}_j)
= \mathbf{u}_0 + \sum_{k=1}^{L}\boldsymbol{\delta}_k(\mathbf{x}_j),
\qquad
\boldsymbol{\delta}_k(\mathbf{x}_j)
= W_{\mathrm{out}}\!\sum_{h} w_{h,k}(\mathbf{x}_j)\,W_O^h W_V^h\tilde{\mathbf{z}}_k,
\label{eq:exact-decomp}
\end{equation}
where $\mathbf{u}_0$ is the decoder output evaluated with all
$z_k = 0$, collecting the affine offsets. Unlike classical
partition-of-unity interpolation, where a fixed field is blended by
the weights, here each anchor synthesizes its own local basis through
$W_V^h\tilde{\mathbf{z}}_k$, and thus both the partition and the basis
are adaptive, conditioned on the input. Similar to this decomposition,
an analogy with classical POD expansion can be made in the
single-headed case as shown in Appendix~\ref{decomposition-appendix}.

\section{Datasets}\label{datasets-section}

To examine Courant's performance, two simulation environments inspired
by industrial problems and a classical cylinder-obstructed flow
environment are considered. These datasets are briefly described below
(detailed setups and visualizations are provided in
Appendix~\ref{datasets-appendix}).

\textbf{2D Cylinder-Obstructed Flow (Cyl.Flow).} It describes the
classic K\'arm\'an vortex street, a repeating pattern of swirling
vortices triggered by flow over a blunt body at a Reynolds number of
approximately 90~\citep{kundu2024fluid}. In this transient simulation,
we model flow over an infinitely long cylinder within a channel formed
by a pair of infinite parallel plates and simulate a single slice. The
training dataset includes cylinders of various sizes and locations.

\textbf{3D Branched Pipe Flow (Br.Pipe).} It models steady-state flow
in a Y-shaped HVAC duct where flow rate in each channel is controlled
by baffle plates. The non-smooth turning angles created by the baffle
plates induce highly turbulent flow at the duct junction. To fully
capture the physical phenomena, the training data includes various
duct geometries, baffle plate angles, and inlet velocities.

\textbf{3D Centrifugal Pump (Ctr.Pump).} It also models a
steady-state flow. It has two main components: an impeller attached to
a rotating shaft and a stationary casing (or volute) enclosing the
impeller. The impeller comprises several curved blades (or vanes)
arranged in a regular pattern around the shaft. As the impeller
rotates, fluid is drawn in through the eye of the casing and flows
radially outward. The rotating blades add energy to the fluid,
increasing both pressure and absolute velocity as the fluid flows from
the eye to the blade periphery. The flow is then discharged through a
diffuser-like channel, resulting in a uniform velocity distribution
with increased pressure at the outlet.

\begin{table}[htb]
\caption{Unstructured mesh benchmarks (NMAE).}
\label{tab:bench-unstr}
\centering
\begin{tabular}{@{}lcccc@{}}
\toprule
& \multicolumn{3}{c}{Steady} & Unsteady \\
\cmidrule(lr){2-4}\cmidrule(l){5-5}
& Br.Pipe & Ctr.Pump & Elast.$^{\dagger}$ & Cyl.Flow \\
\midrule
ENF        & ---                 & ---                  & ---                  & \sn{2.3}{1} \\
MGN        & \sn{2.2}{1}         & OOM                  & ---                  & \first{\sn{1.3}{1}} \\
Transolver & \first{\sn{8.6}{2}} & \second{\sn{2.9}{1}} & \first{\sn{5.7}{3}}  & \first{\sn{1.3}{1}} \\
UPT        & \sn{3.1}{1}         & \sn{4.1}{1}          & \second{\sn{1.3}{2}} & \sn{2.8}{1} \\
\midrule
Courant    & \second{\sn{9.6}{2}} & \first{\sn{1.7}{1}} & \sn{2.1}{2}          & \second{\sn{1.7}{1}} \\
\bottomrule
\end{tabular}

\medskip
{\footnotesize --- = architecture incompatible with dataset.\quad
$^{\dagger}$Elasticity is included for cross-comparison; the other
three columns are datasets described in \S\ref{datasets-section}}
\end{table}

\begin{table}[htb]
\caption{Structured grid benchmarks (NMAE).}
\label{tab:bench-str}
\centering
\begin{tabular}{@{}lccccc@{}}
\toprule
& \multicolumn{3}{c}{Steady} & \multicolumn{2}{c}{Unsteady} \\
\cmidrule(lr){2-4}\cmidrule(l){5-6}
& Darcy & Airfoil & Pipe & Diff-Sorp & NS-3D \\
\midrule
ENF        & ---                  & ---                  & ---                  & \sn{4.9}{3}          & \sn{3.8}{2} \\
FNO        & \sn{8.5}{3}          & \second{\sn{2.1}{3}} & \sn{2.2}{3}          & \second{\sn{3.9}{3}} & \sn{1.9}{2} \\
FactFormer & \first{\sn{4.2}{3}}  & \sn{3.2}{3}          & \first{\sn{1.3}{3}}  & ---                  & --- \\
Transolver & \second{\sn{4.9}{3}} & \first{\sn{1.9}{3}}  & \second{\sn{1.8}{3}} & \second{\sn{3.9}{3}} & \second{\sn{1.8}{2}} \\
UPT        & \sn{1.6}{2}          & \sn{5.0}{2}          & \sn{5.9}{3}          & \sn{1.9}{2}          & \first{\sn{1.7}{2}} \\
\midrule
Courant    & \sn{5.7}{3}          & \sn{1.4}{2}          & \sn{1.4}{2}          & \first{\sn{3.8}{3}}  & \first{\sn{1.7}{2}} \\
\bottomrule
\end{tabular}

\medskip
{\footnotesize --- = architecture incompatible with dataset.}
\end{table}

\section{Results and Discussion}\label{sec:resultsanddiscussion}

We evaluate Courant along two axes: predictive accuracy on benchmarks
and industrially-motivated simulations, and the interpretability of
the learned latent representations from both spatial and temporal
perspective.

\begin{figure}[htb]
    \centering
    \begin{minipage}[t]{0.45\linewidth}
        \begin{subfigure}[t]{\linewidth}
            \centering
            \includegraphics[width=\linewidth]{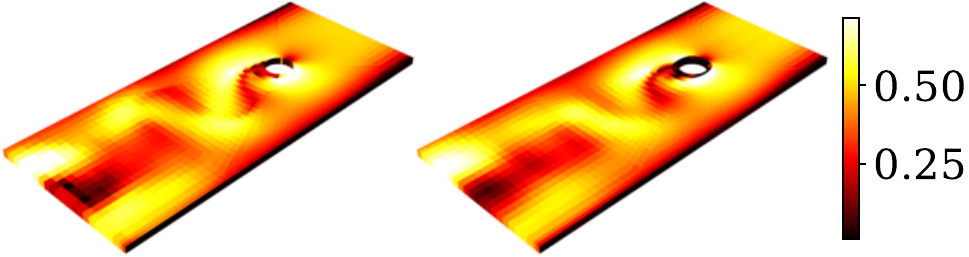}
            \caption{2D Cylinder-Obstructed Flow}
            \label{fig:pred-cf}
        \end{subfigure}

        \vspace{0.5em}

        \begin{subfigure}[t]{\linewidth}
            \centering
            \includegraphics[width=0.9\linewidth]{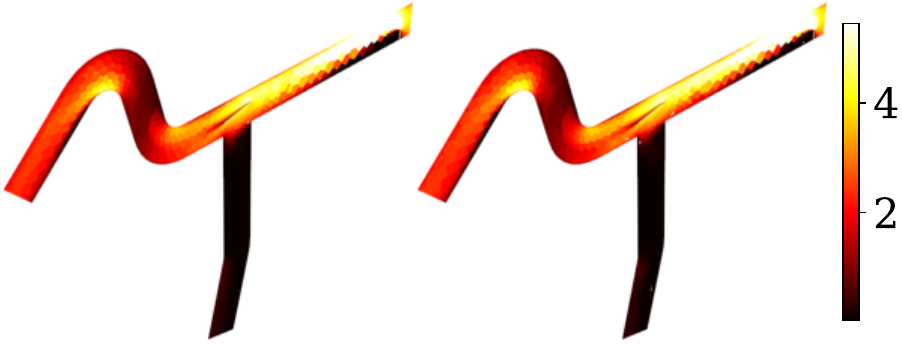}
            \caption{3D Branched Pipe Flow}
            \label{fig:pipe-gw-s2-truepred}
        \end{subfigure}
    \end{minipage}
    \hspace{0.02\linewidth}
    \begin{minipage}[t]{0.45\linewidth}
        \vspace{-35pt}
        \begin{subfigure}[t]{\linewidth}
            \centering
            \includegraphics[width=\linewidth]{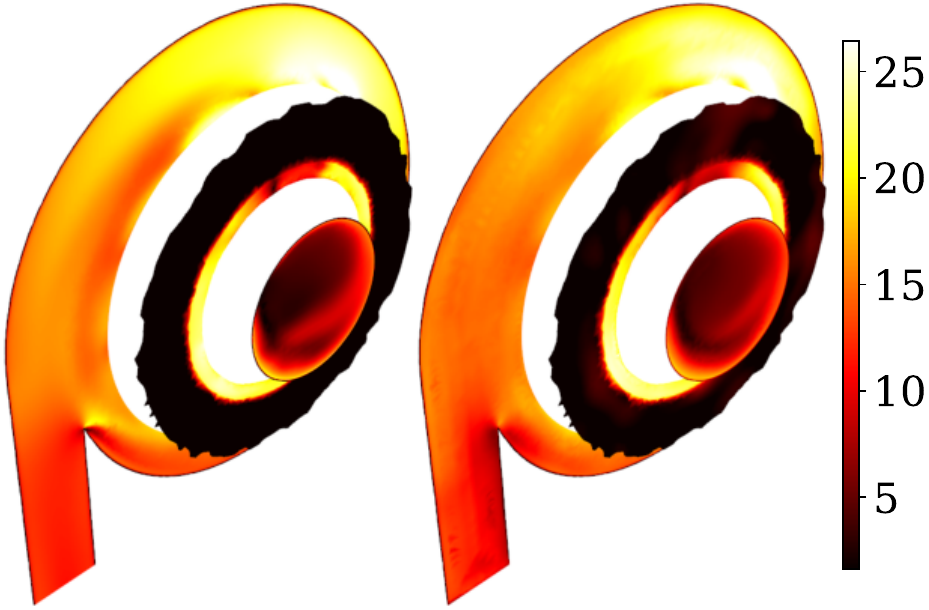}
            \caption{3D Centrifugal Pump}
            \label{fig:pred-pump}
        \end{subfigure}
    \end{minipage}
    \caption{Ground truth (left) and Courant prediction (right) for
    representative test samples showing velocity norm. For 2D
    Cylinder-Obstructed Flow the final time step ($t=10$) of a prediction
    is shown. Visualization of 3D Branched Pipe Flow is available in
    Appendix~\ref{sec:3d-branched-pipe-visualizations}.}
    \label{fig:predictions}
\end{figure}

\subsection{Predictive accuracy}\label{sec:predictiveaccuracy}

\paragraph{Industrially motivated benchmarks}
Table~\ref{tab:bench-unstr} reports normalized mean absolute error
(NMAE) on our three industrial benchmarks --- Cyl.Flow, Br.Pipe, and
Ctr.Pump --- alongside the Elasticity dataset borrowed from the Neural
Solver Library for cross-comparison. A representative test prediction
is shown in Figure~\ref{fig:predictions} and additional
visualizations of ground truth and predictions from the test set are
provided in Appendix~\ref{appendix-decompositions}. Courant is
competitive with the strongest baselines on the smaller datasets
(Cyl.Flow, Br.Pipe) and outperforms the Transolver baseline in these
experiments on the large dataset, Ctr.Pump, while training with
substantially fewer parameters and lower peak memory
(Appendix~\ref{sec:resource-efficiency}).

These gains are driven by Courant's architectural choices, validated
by the per-component ablation in Appendix~\ref{ablation-appendix}.
Geometry-anchored latents with the shared encoder/decoder coordinate
embedding, the self-attention NeuralODE processor on the transient
case, design-parameter conditioning at the encoder, and
distance-field aware decoder queries each contribute to the predictive
accuracy reported above. Beyond accuracy, the decoupled input--output
design gives Courant a structural advantage on steady problems: the
encoder can ingest just the boundary point cloud while the decoder
evaluates at arbitrary query coordinates (\S\ref{sec:decoder}). This
preserves accuracy on Br.Pipe and Ctr.Pump while reducing training
time and peak memory by $2$--$4\times$
(Appendix~\ref{sec:encoder-bpc}). UPT~\citep{alkin2024universal}
adopts a closely related decoupled input--output Perceiver-style
design but typically processes the full input mesh; Courant
outperforms UPT on every industrial benchmark.

\paragraph{Additional benchmarks}\label{sec:additionalbenchmarks}
To position Courant against well-known surrogate models, we further
evaluate on standardized datasets and baselines from the Neural Solver
Library~\citep{wu2024transolver}, spanning 1D, 2D, and 3D problems in
both steady and unsteady
regimes.\footnote{The original repository code was retrieved from
\url{https://github.com/thuml/Neural-Solver-Library}.} We compare
against the provided implementations of FNO~\citep{li2021fourier},
Transolver~\citep{wu2024transolver},
FactFormer~\citep{li2023scalable}, and
UPT~\citep{alkin2024universal}, and add our own implementation of
ENF~\citep{wessels2025enf} on the unsteady datasets.

Results in Table~\ref{tab:bench-str} mirror the pattern seen on the
industrial benchmarks. Methods specialized for structured grids, FNO
and FactFormer, achieve the lowest error on the small
structured-grid steady tasks. Transolver, although mesh-agnostic,
also performs strongly across both regimes and is the best on Elast.\
and Airfoil. Courant trails on the small structured tasks --- an
expected gap given its decoupled input--output Perceiver design ---
but is competitive on the larger benchmarks and ties UPT for the best
score on NS-3D. Mirroring the industrial trend, Courant outperforms
UPT on most of these benchmarks.

\subsection{Latent space interpretability}\label{sec:latentspaceinterpretability}

We now turn to the structured latent representations that adaptively
develop under end-to-end training. The architecture is designed to
support such structure through its inductive biases, but the specific
representations that emerge are not directly supervised. We analyze
them along two dimensions: spatial localization/basis decomposition
and latent dynamics.

\paragraph{Spatial localization of basis-like decomposition}

\begin{figure}[htb]
\centering
\includegraphics[width=1.0\linewidth]{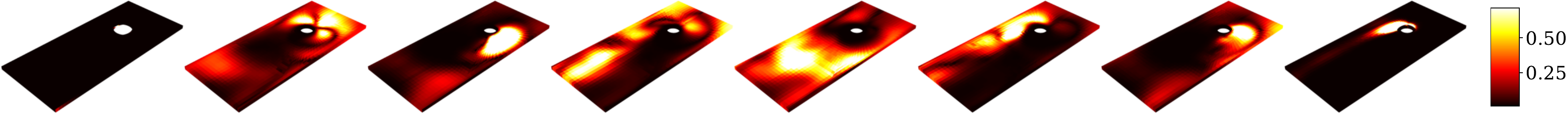}
\caption{Top latent decompositions $\boldsymbol{\delta}_k$ for the 2D
Cylinder-Obstructed Flow with highest norm field. Latents are ranked
left to right.}\label{decomposition-cf-figure}
\end{figure}

\begin{figure}[htb]
    \centering
\includegraphics[width=1.0\linewidth]{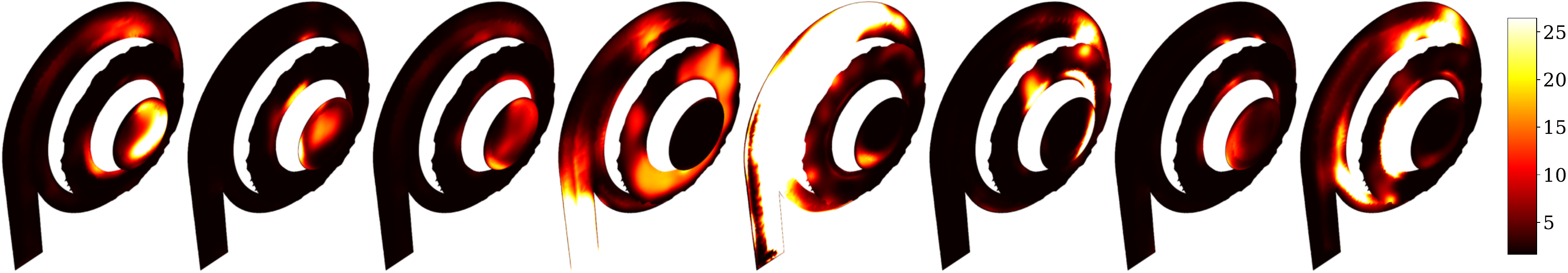}
\includegraphics[width=1.0\linewidth]{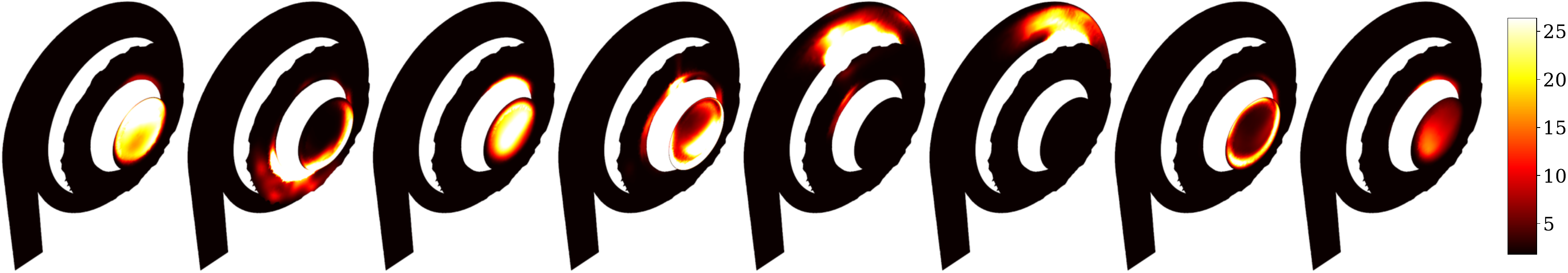}
\caption{Top 8 latent decompositions $\boldsymbol{\delta}_k$ with
highest norm velocity field for the centrifugal pump (\textbf{top})
and with added GWA bias with details provided in
Appendix~\ref{sec:gwa} (\textbf{bottom}). Decoded pressure for each
latent is visualized at three evenly spaced cross
sections.}\label{pump-flow-modes-figure}
\end{figure}

Figure~\ref{decomposition-cf-figure} shows the decomposed field from
the top 8 anchors with highest norm or highest peak for the 2D
Cylinder-Obstructed Flow. The decomposed field of the latents is
predominantly spatially compact and mutually disjoint: the top-norm
latent anchors cleanly separate into distinct flow regions tied to the
cylinder geometry and wake structure. The 3D Branched Pipe
(Appendix~\ref{sec:3d-branched-pipe-visualizations}) and 3D
Centrifugal Pump (Fig.~\ref{pump-flow-modes-figure}) datasets show a
similar pattern, but these steady datasets are qualitatively less
compact. Additional basis decompositions are available for each
dataset in Appendix~\ref{appendix-decompositions}.

To further compact and anchor the decomposable basis in steady
datasets, we trained models with an added Gaussian window adaptive
bias to cross attention layers as detailed in
Appendix~\ref{sec:gwa}. This resulted in comparable model performance
(Appendix~\ref{ablation-appendix}) with a more local basis as observed
in Figure~\ref{pump-flow-modes-figure} for the 3D Centrifugal Pump.
Since this bias appears to be successful at compacting the decomposed
basis derived from steady datasets, future study is warranted towards
its application in the time-dependent regime.

\paragraph{Latent space dynamic analysis}

\begin{figure}[htb]
    \centering
    \includegraphics[width=1\linewidth]{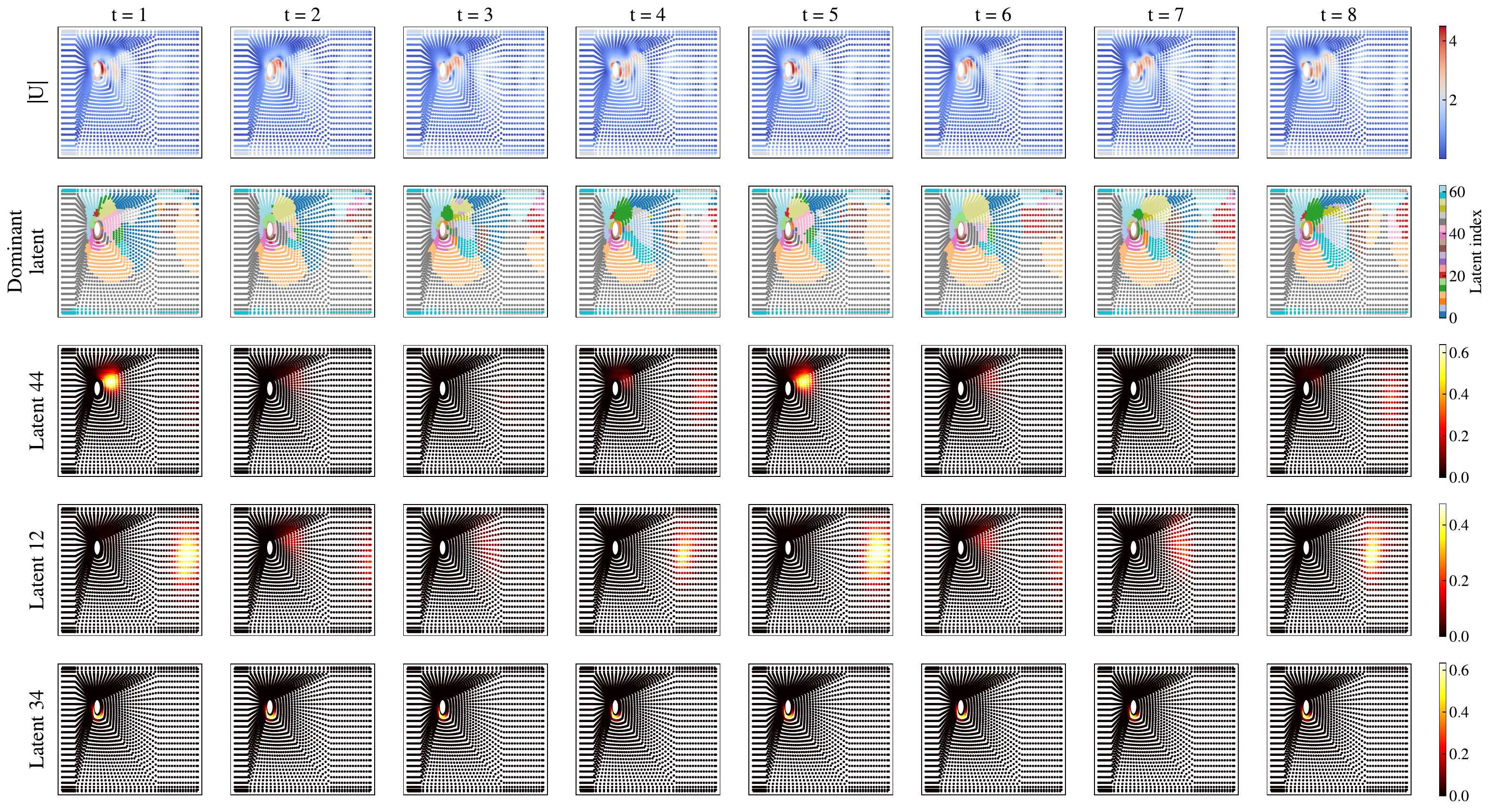}
    \caption{Spatial localization and latent anchor specialization on
    the 2D Cylinder-Obstructed Flow. Top row: predicted velocity norm
    $|u|$ over time. Second row: dominant-latent partitioning of the
    domain at each time step. Bottom three rows: decoder attention
    weights $w_{\cdot,k}(\mathbf{x})$ for three representative
    anchors --- the top two are dynamic modes, the bottom one is
    stationary.}
    \label{fig:cf_temporal_variation}
\end{figure}

We now turn to a detailed dynamical analysis of Courant trained with
the 2D Cylinder-Obstructed Flow dataset.
Figure~\ref{fig:cf_temporal_variation} reveals that the latent space
organizes spatially in a way that is tightly coupled to the physical
flow. The dominant-latent map in (row~2) categorizes each point by
the highest attention-weighted latent anchor. We find that it
partitions the domain into compact, contiguous regions, each governed
by a single latent anchor. The per-anchor decoder attention weights
(rows~3--5) reveal three distinct physical roles: the first map tracks
vortex shedding in the near wake, just downstream of the cylinder; the
second captures the same shedding pattern further downstream in the
far wake; and the third remains pinned to the cylinder's boundary
layer as a stationary mode.

\begin{figure}[htb]
    \centering
    \begin{subfigure}[t]{0.45\linewidth}
        \centering
        \includegraphics[width=\linewidth]{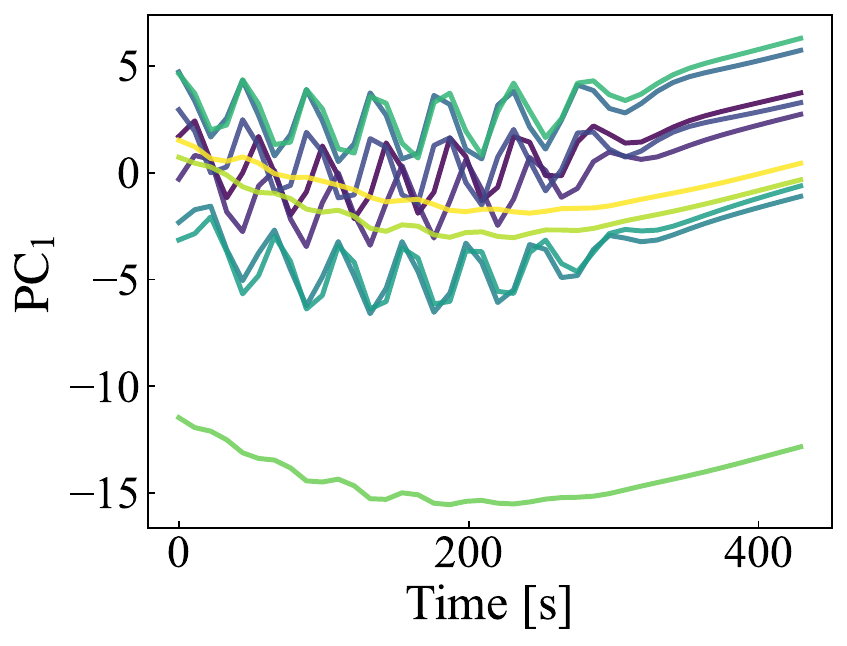}
        \caption{PC1 latent time series.}
        \label{fig:psd_pc1_ts}
    \end{subfigure}
    \hspace{0.05\linewidth}
    \begin{subfigure}[t]{0.42\linewidth}
        \centering
        \includegraphics[width=\linewidth]{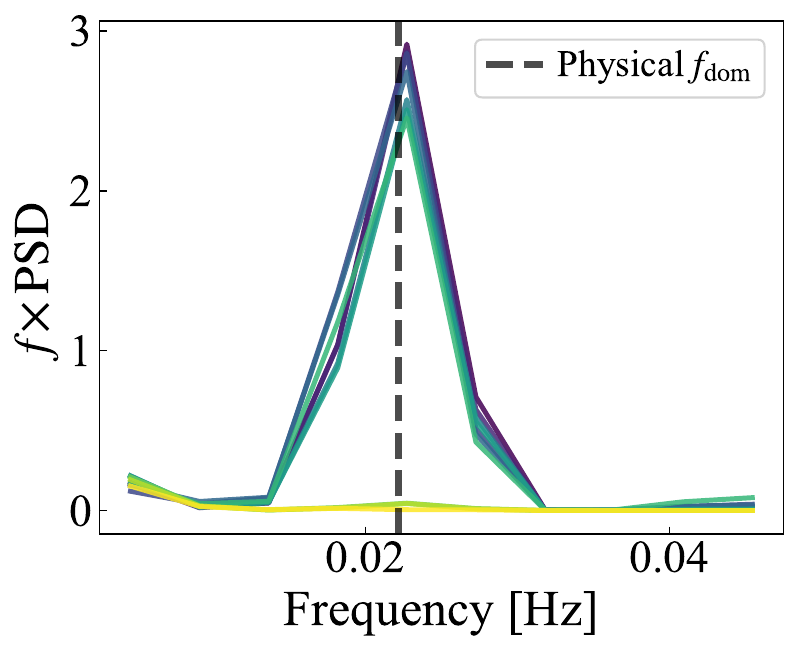}
        \caption{PC1 premultiplied PSD.}
        \label{fig:psd_pc1_psd}
    \end{subfigure}

    \vspace{0.75em}

    \begin{subfigure}[t]{0.44\linewidth}
        \centering
        \includegraphics[width=\linewidth]{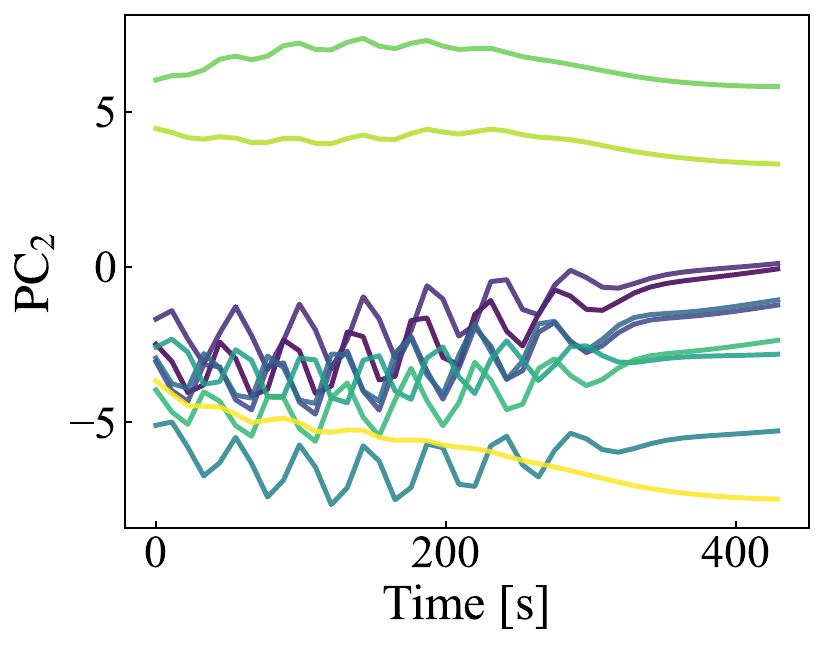}
        \caption{PC2 latent time series.}
        \label{fig:psd_pc2_ts}
    \end{subfigure}
    \hspace{0.02\linewidth}
    \begin{subfigure}[t]{0.44\linewidth}
        \centering
        \includegraphics[width=\linewidth]{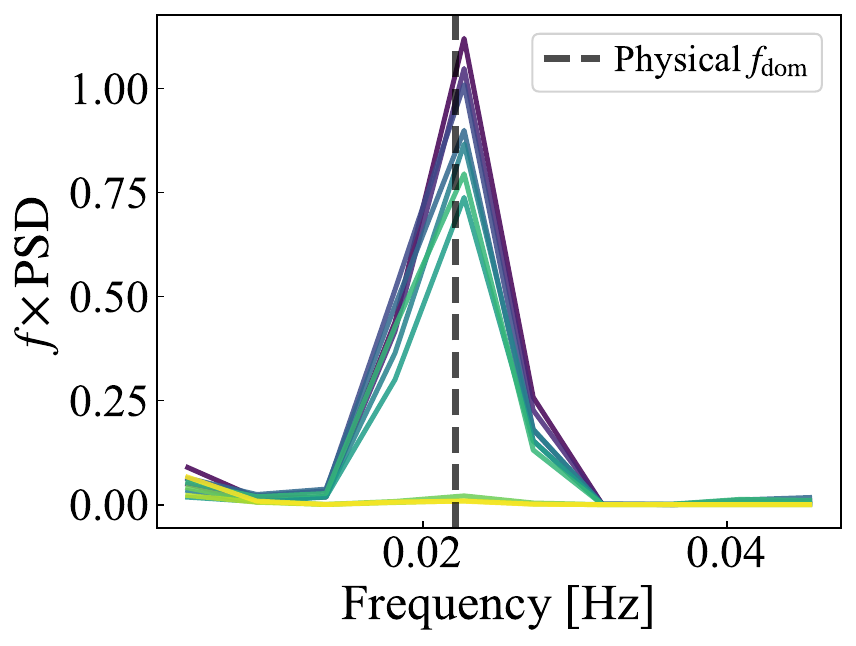}
        \caption{PC2 premultiplied PSD.}
        \label{fig:psd_pc2_psd}
    \end{subfigure}

    \caption{%
        ODE-evolved latent trajectories projected onto the leading two
        principal components (PC1, top row; PC2, bottom row) across
        representative latent anchors.
        Left column: PC time series over the simulation window.
        Right column: premultiplied Welch PSD $f\!\cdot\!\mathrm{PSD}$;
        the dashed line marks the physical wake-probe shedding frequency
        $f_{\text{shed}}\!\approx\!0.022\,\text{Hz}$.
        The spectral peak aligns with $f_{\text{shed}}$ in both components,
        indicating that the latent dynamics have internalized the dominant
        vortex-shedding frequency.
        See Appendix~\ref{vortex-shedding-section} for methodology.%
    }
    \label{fig:cf_latent_psd}
\end{figure}

The K\'arm\'an vortex street imposes a single dominant timescale on
this flow: at $\mathrm{Re}\!\approx\!90$, the cylinder wake sheds
vortices at $f_{\text{shed}}\!\approx\!0.022$, recovered cleanly from
a transverse-velocity probe in the wake
(Fig.~\ref{fig:cf_shedding_frequency}).
Figure~\ref{fig:cf_latent_psd} applies the same diagnostic on the
latent side: a Welch Power Spectral Density (PSD) of the leading principal components of the
NeuralODE-evolved latents yields a sharply peaked spectrum whose
maximum coincides with the physical $f_{\text{shed}}$, indicating
that the NeuralODE has learned to advance the latent state at the
correct physical timescale.

\begin{figure}[htb]
    \centering
    \begin{subfigure}[t]{0.55\linewidth}
        \centering
        \includegraphics[width=\linewidth]{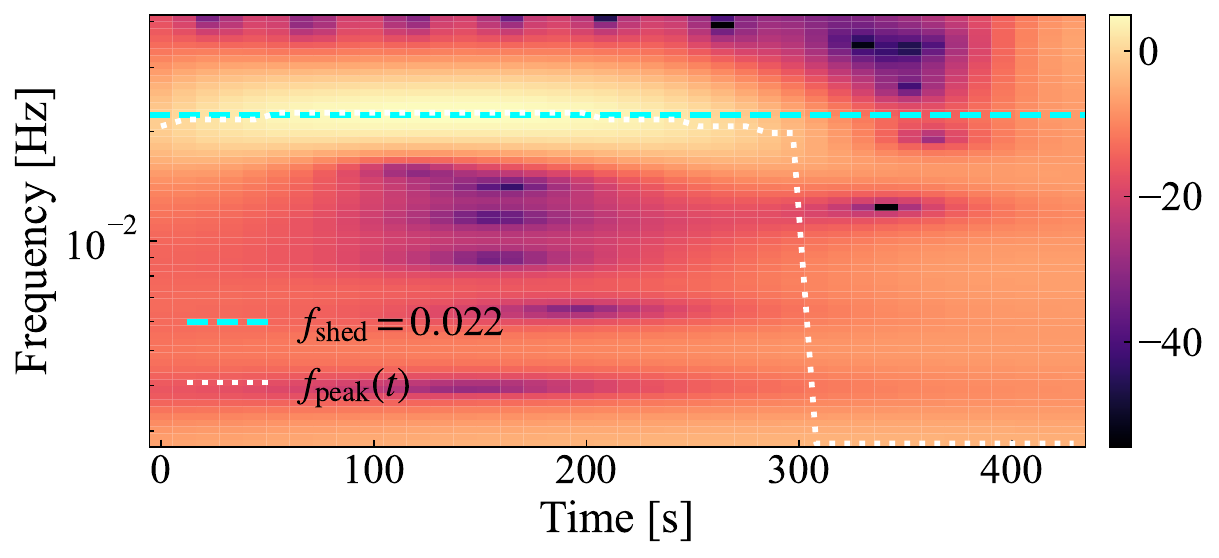}
        \caption{Morlet scalogram (dB).}
        \label{fig:cf_scalogram}
    \end{subfigure}
    \hfill
    \begin{subfigure}[t]{0.37\linewidth}
        \centering
        \includegraphics[width=\linewidth]{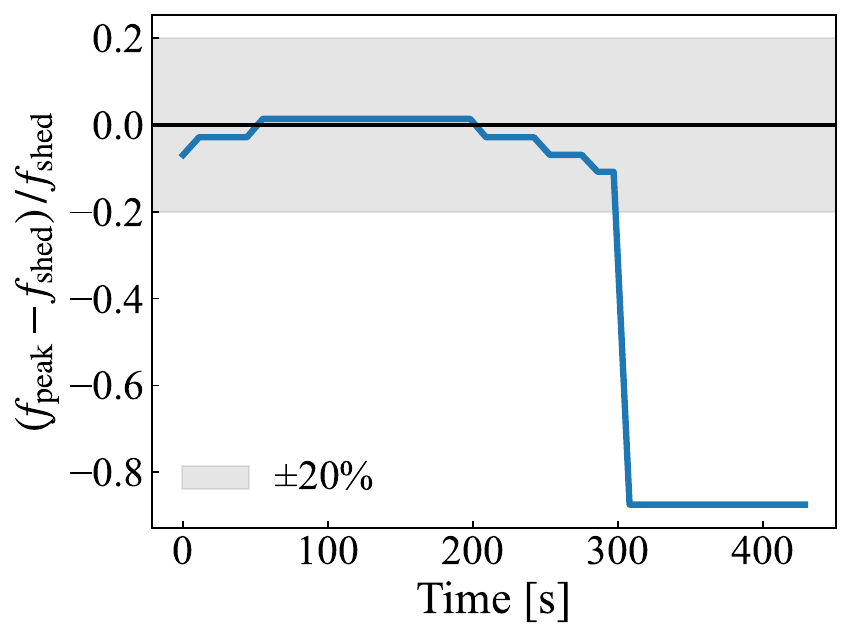}
        \caption{Relative peak-frequency drift.}
        \label{fig:cf_peak_drift}
    \end{subfigure}
    \caption{Time-resolved frequency content of a representative latent
    anchor's leading PC over the ODE rollout.
    (\subref{fig:cf_scalogram}) complex Morlet wavelet scalogram with the
    physical shedding frequency $f_{\text{shed}}=0.022\,\text{Hz}$
    overlaid (dashed) and the per-time spectral peak $f_{\text{peak}}(t)$
    tracked (dotted); strong energy concentration at $f_{\text{shed}}$
    persists throughout the rollout.
    (\subref{fig:cf_peak_drift}) relative drift
    $(f_{\text{peak}}-f_{\text{shed}})/f_{\text{shed}}$
    with a $\pm20\%$ reference band, quantifying the stability of that
    lock-in.
    Methodology in Appendix~\ref{vortex-shedding-section}.}
    \label{fig:cf_latent_scalogram}
\end{figure}

Time--frequency content reveals when this match degrades.
Figure~\ref{fig:cf_latent_scalogram} shows a complex Morlet wavelet
scalogram of a representative latent's leading PC (left) and the
relative drift of its instantaneous peak frequency
$f_{\text{peak}}(t)$ from $f_{\text{shed}}$ (right). The shedding
peak is preserved well past the training horizon ($\!\sim\!100$~s),
with $f_{\text{peak}}$ remaining within $\pm 20\%$ of $f_{\text{shed}}$
for roughly $300$~s; beyond that the peak slides to lower frequencies,
signalling that the latent state has left the limit cycle. Because the
diagnostic operates entirely on the latent state, it is a candidate
self-monitoring signal at deployment time. The same loss of
oscillatory content is visible in the eigenspectrum of the NeuralODE
Jacobian (Appendix~\ref{sec:eigenspectrum-appendix}).

\section{Conclusion}\label{sec:conclusion}

We have presented an encoder--processor--decoder surrogate where
interpretable latent decomposition is the primary design objective.
The architecture combines geometry-anchored latent queries with a
shared random Fourier coordinate embedding in a Perceiver encoder, a
self-attention NeuralODE processor, and a lightweight decoder. These
architectural choices also improve predictive accuracy and keep
Courant competitive across both industrial and standardized
benchmarks; the decoupled input--output design lets the encoder
operate on information-dense inputs such as boundary point clouds
while the decoder can evaluate a separate, larger point cloud.
Contrary to domain decomposition as a means to enable neural
sparsity, these architecture choices enable the scheme to function
like a locally adaptive \textit{hp}-refinement scheme where local
adaptive basis functions are discovered to cohere with the multiscale
features of the solution, while their local support is strongly
promoted.

Across three examined datasets, the latent representations develop
physically meaningful structure under end-to-end training with only
an output prediction loss. Latent anchors partition the simulation
domain into compact, distinct regions; within a single rollout, some
latent anchors track transient features (dynamic modes) while others
remain anchored to the geometry (stationary modes). The per-anchor
decoder contributions form an exact additive decomposition of the
predicted field, yielding a state-dependent, partition-of-unity-like
spatial basis rather than the fixed, linear basis of classical
reduced-order representations such as POD. With appropriate
architectural choices, the internal representations of a neural
surrogate can be made legible to domain practitioners: per-anchor
decoder contributions can be visualized as compact basis functions,
the latent PSD and per-anchor scalograms can
expose the dynamical content of the rollout. Together these signals
support practical use cases including in-deployment monitoring of
rollout quality, without sacrificing the expressivity that motivates
neural approaches over classical linear reduced-order methods.

\paragraph{Limitations}
First, the interpretability analyses presented here are primarily
qualitative and visual; the spatial decomposition and temporal
coherence claims rely on inspection rather than formal metrics, which
limits their generality. Second, on benchmarks with regular structured
grids, grid-native methods that exploit spectral structure outperform
our mesh-agnostic architecture; the generality of the Perceiver
encoder comes at a cost on problems where that generality is not
needed. Third, benchmark coverage for unstructured and industrially
complex geometries remains limited to two time-independent datasets;
broader evaluation across a wider range of physics, geometric
complexity, and scale would strengthen the claims.

\paragraph{Acknowledgements}
The authors acknowledge the contributions of Diego Andrade and Jau-Wei Chen for industrially-motivated simulation dataset preparation.


\PLrefheading
\bibliography{references}

\newpage
\appendix

\section{Datasets and Simulations}\label{datasets-appendix}

The simulation configurations used in this paper are summarized in
Table~\ref{tab:sim-setup}. All simulations are solved using the
Finite Volume Method (FVM) with different software packages
(i.e., OpenFoam v2206~\citep{openfoam} and Ansys Fluent
(2024R1)~\citep{ansysfluent}). A laminar flow model is used for the
2D Cylinder-Obstructed Flow to capture the low Reynolds number
behavior, while a Reynolds-Averaged Navier-Stokes (RANS) turbulence
model is used for the other two cases to capture turbulent behavior.
A velocity-inlet and pressure-outlet boundary condition pair is
applied to all cases. For the 2D Cylinder-Obstructed Flow, the inlet
velocity ramps from 0~m/s to a parabolic profile between t=0~s and
t=1~s, then remains constant thereafter. For the 3D Branched Pipe
Flow, a 1/7th power-law velocity profile is applied at the inlet.
For the 3D Centrifugal Pump, a uniform velocity is enforced at the
inlet. The Moving Reference Frame (MRF) approach
\citep{batchelor2000introduction} is applied to model the rotating
effect in the centrifugal pump; otherwise, the simulations are typical
internal flow cases. The 2D Cylinder-Obstructed Flow is a transient
simulation with a time step size of 0.1~s and a simulation duration
from 0 to 1000~s, while the other two simulations model steady-state
phenomena and employ an adaptive time-stepping strategy.

In addition to the simulation configuration,
Table~\ref{tab:sim-design} summarizes the design parameters
(variables) used to generate the training dataset for each simulation.

\begin{table}[]
\caption{Summary of simulation configurations.}
\label{tab:sim-setup}
\centering
\begin{tabular}{c|ccc}
\hline
 & Cyl.Flow & Br.Pipe & Ctr.Pump \\
\hline
cell number & $O(2e3)$ & $O(3.7e4)$ & $O(3e5)$ \\
\hline
fluid model & air & air & water \\
\hline
turbulent model & laminar & RANS & RANS \\
\hline
\shortstack[c]{boundary\\conditions}
  & \multicolumn{3}{c}{\shortstack[c]{velocity inlet\\pressure outlet\\no-slip wall}} \\
\hline
time step size & 0.1 (s) & adaptive & adaptive \\
\hline
solver
  & \shortstack[c]{OpenFOAM\\FVM}
  & \shortstack[c]{Fluent\\FVM}
  & \shortstack[c]{Fluent\\FVM} \\
\hline
special treatment & - & - & MRF \\
\hline
\end{tabular}
\end{table}

\begin{table}[]
\caption{Design parameters used to generate the simulation datasets,
with the range each parameter is sampled over.}
\label{tab:sim-design}
\centering
\small
\begin{tabular}{lll}
\toprule
Dataset & Parameter & Range \\
\midrule
Cyl.Flow & radius of the cylinder            & [0.75 - 4]~(m) \\
         & center of the cylinder            & $X_{center}$: [-8.0, 8.0]~(m) \\
         &                                   & $Y_{center}$: [-5.0, 5.0]~(m) \\
\midrule
Br.Pipe  & angle of middle baffle plate          & [0 - 60]~(deg)\\
         & aperture length of inlet baffle plate & [20 - 37]~(mm)\\
         & angle between ducts                   & [75 - 120]~(deg) \\
         & curvature angle of duct branch        & [90 - 180]~(deg)\\
         & inlet mean velocity                   & [3 - 10]~(m/s)\\
\midrule
Ctr.Pump & inlet radius                          & [160 - 200]~(mm) \\
         & impeller radius                       & [80 - 100]~(mm) \\
         & blade pitch angle                     & [5 - 40]~(deg) \\
         & inlet flow velocity                   & [1.5 - 3.5]~(m/s) \\
\bottomrule
\end{tabular}
\end{table}

\begin{figure}[h]
\centering
\begin{subfigure}{0.30\textwidth}
    \centering
    \includegraphics[width=0.9\linewidth]{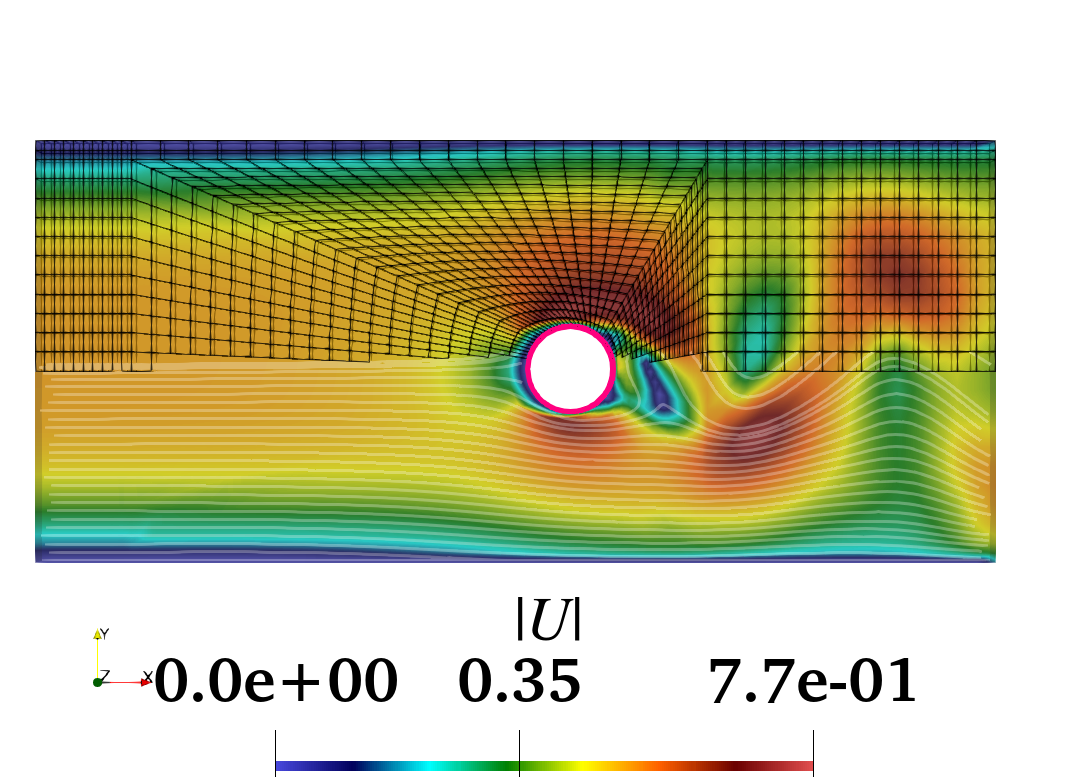}
    \caption{}\label{cylinder-0}
\end{subfigure}%
\begin{subfigure}{0.30\textwidth}
    \centering
    \includegraphics[width=0.9\linewidth]{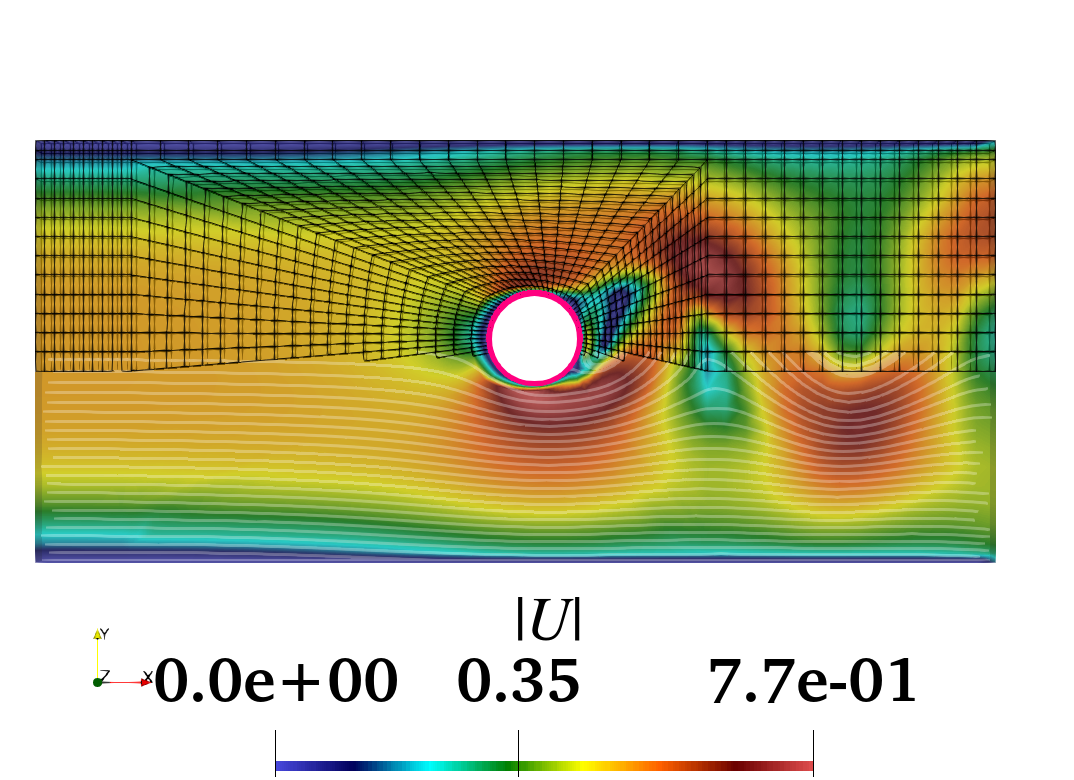}
    \caption{}\label{cylinder-1}
\end{subfigure}%
\begin{subfigure}{0.30\textwidth}
    \centering
    \includegraphics[width=0.9\linewidth]{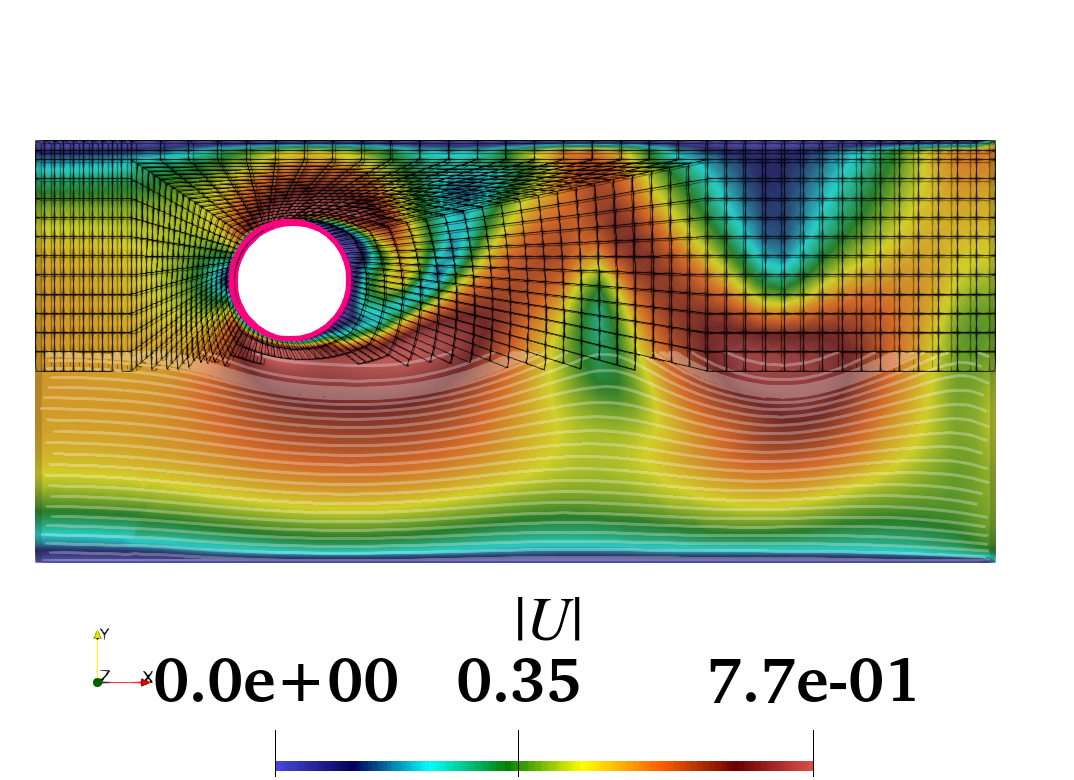}
    \caption{}\label{cylinder-2}
\end{subfigure}

\vspace{0.5em}

\begin{subfigure}{0.30\textwidth}
    \centering
    \includegraphics[width=0.9\linewidth]{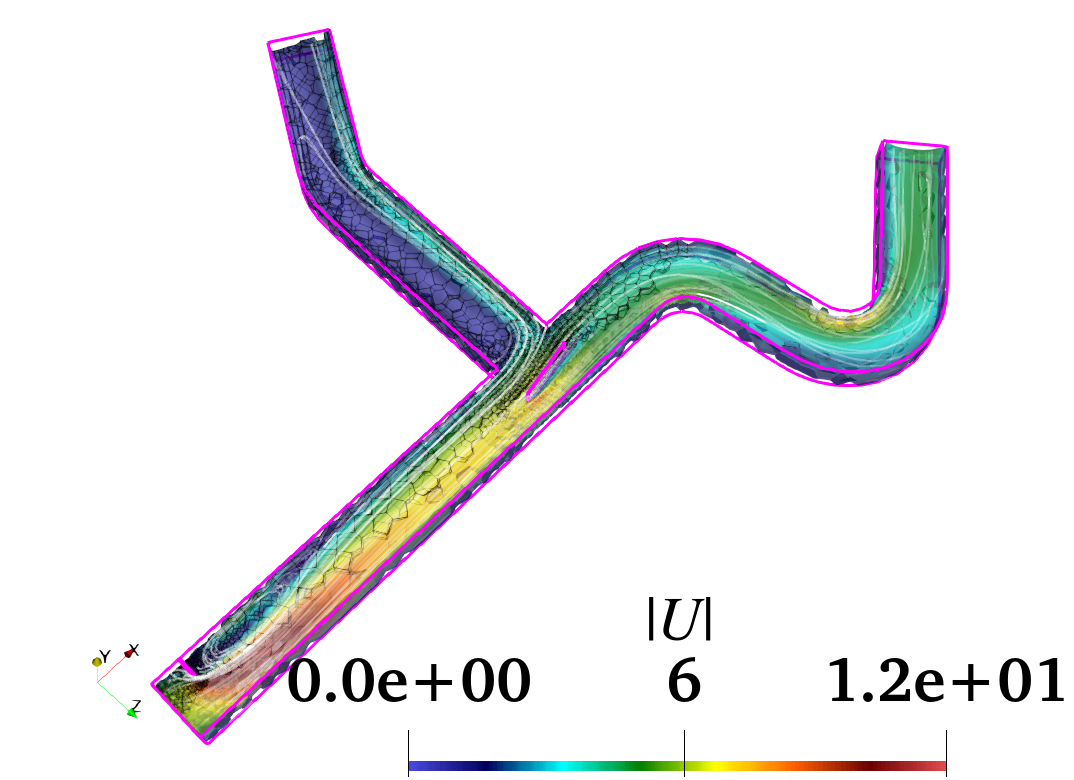}
    \caption{}\label{hvac-0}
\end{subfigure}%
\begin{subfigure}{0.30\textwidth}
    \centering
    \includegraphics[width=0.9\linewidth]{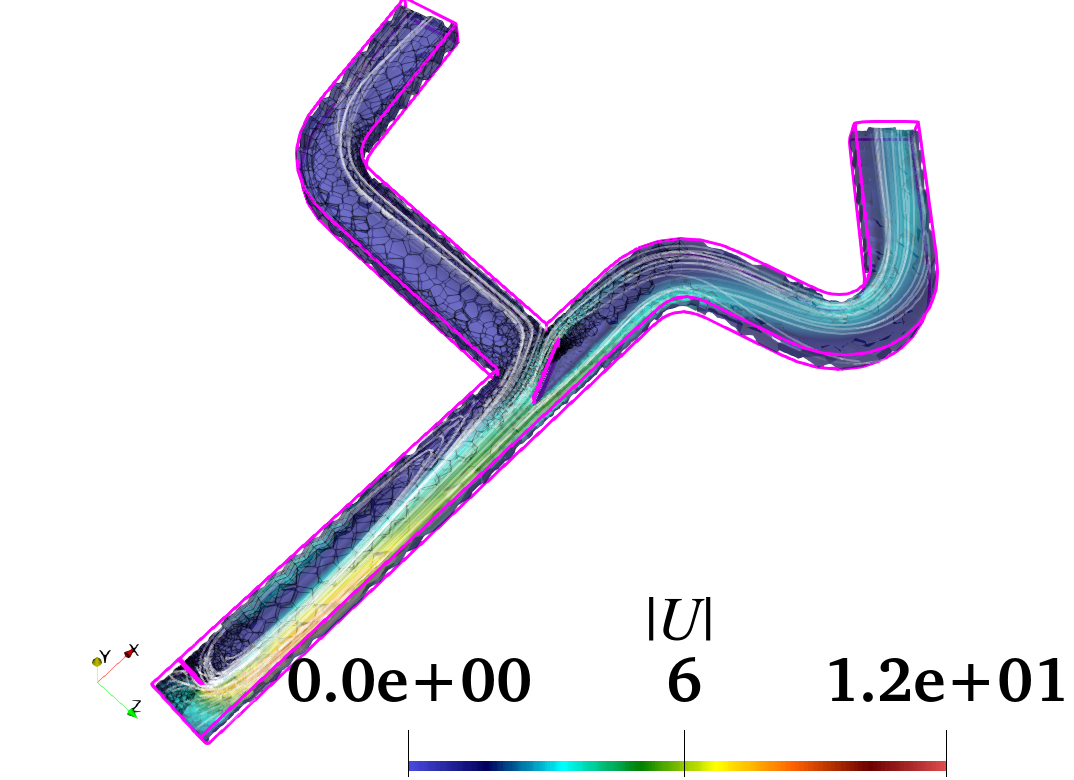}
    \caption{}\label{hvac-1}
\end{subfigure}%
\begin{subfigure}{0.30\textwidth}
    \centering
    \includegraphics[width=0.9\linewidth]{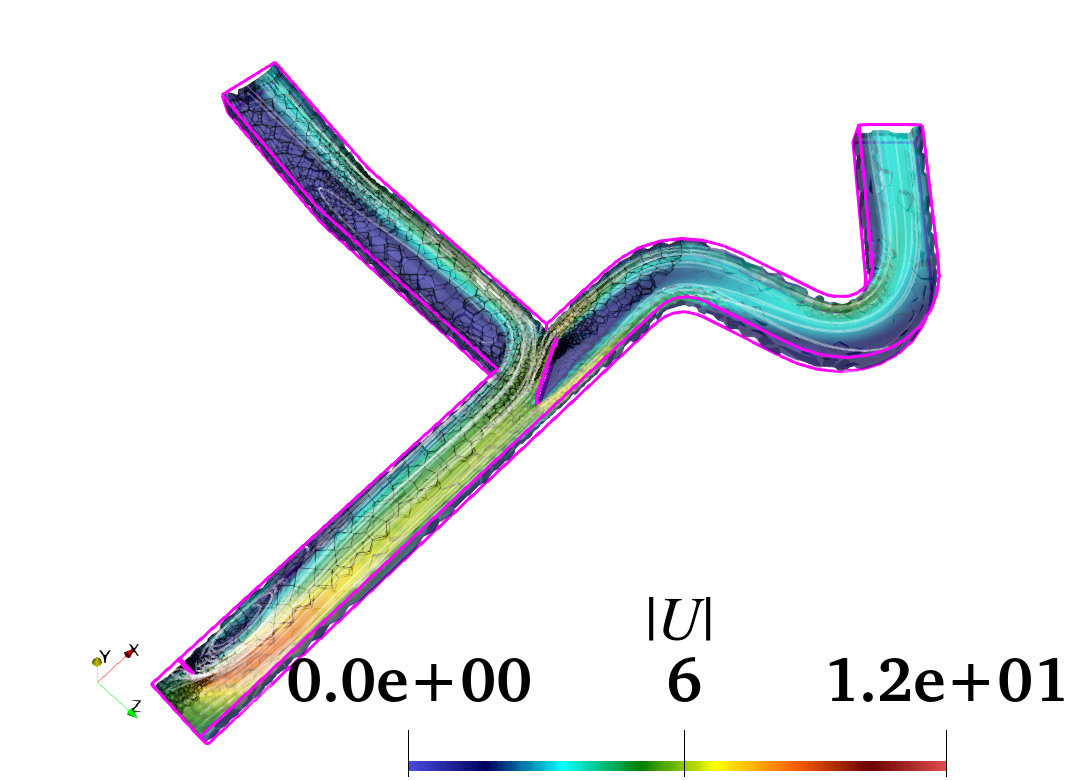}
    \caption{}\label{hvac-2}
\end{subfigure}

\vspace{0.5em}

\begin{subfigure}{0.30\textwidth}
    \centering
    \includegraphics[width=0.9\linewidth]{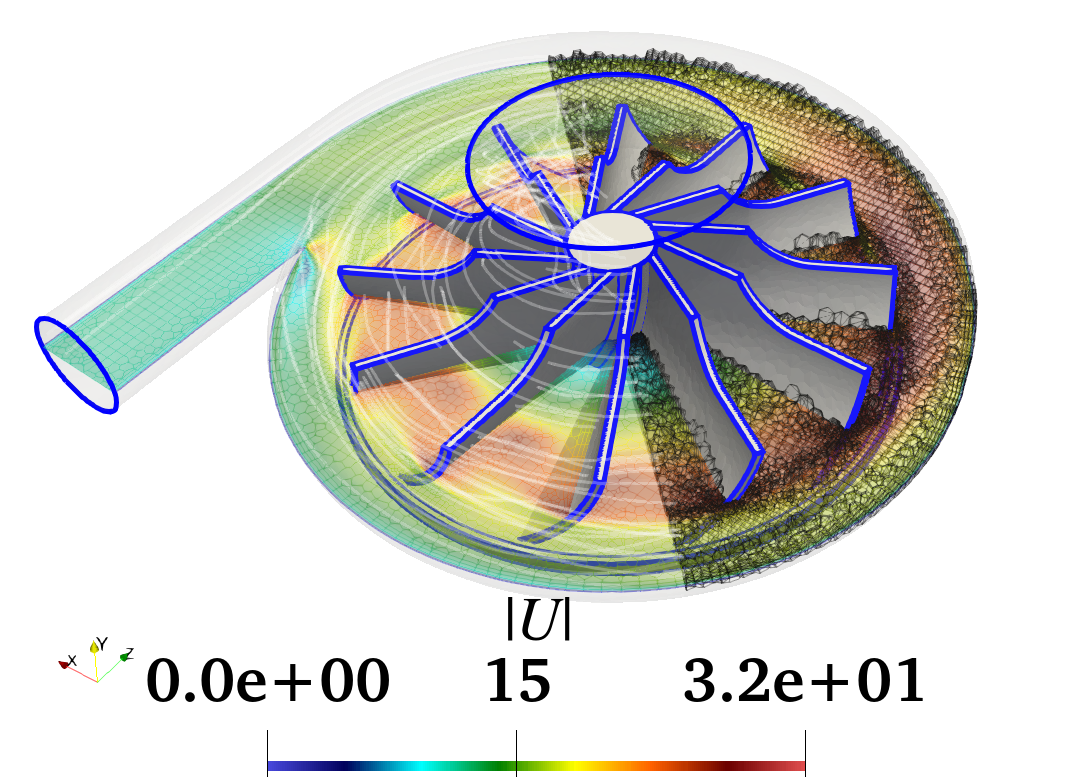}
    \caption{}\label{pump-0}
\end{subfigure}%
\begin{subfigure}{0.30\textwidth}
    \centering
    \includegraphics[width=0.9\linewidth]{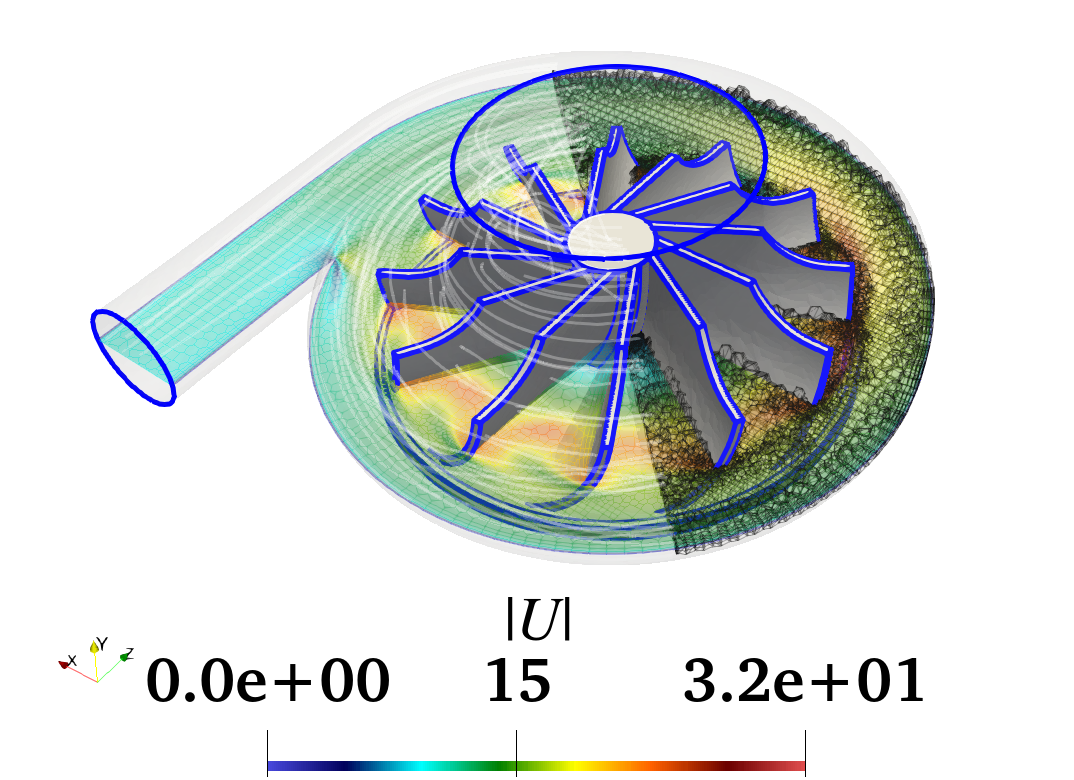}
    \caption{}\label{pump-1}
\end{subfigure}%
\begin{subfigure}{0.30\textwidth}
    \centering
    \includegraphics[width=0.9\linewidth]{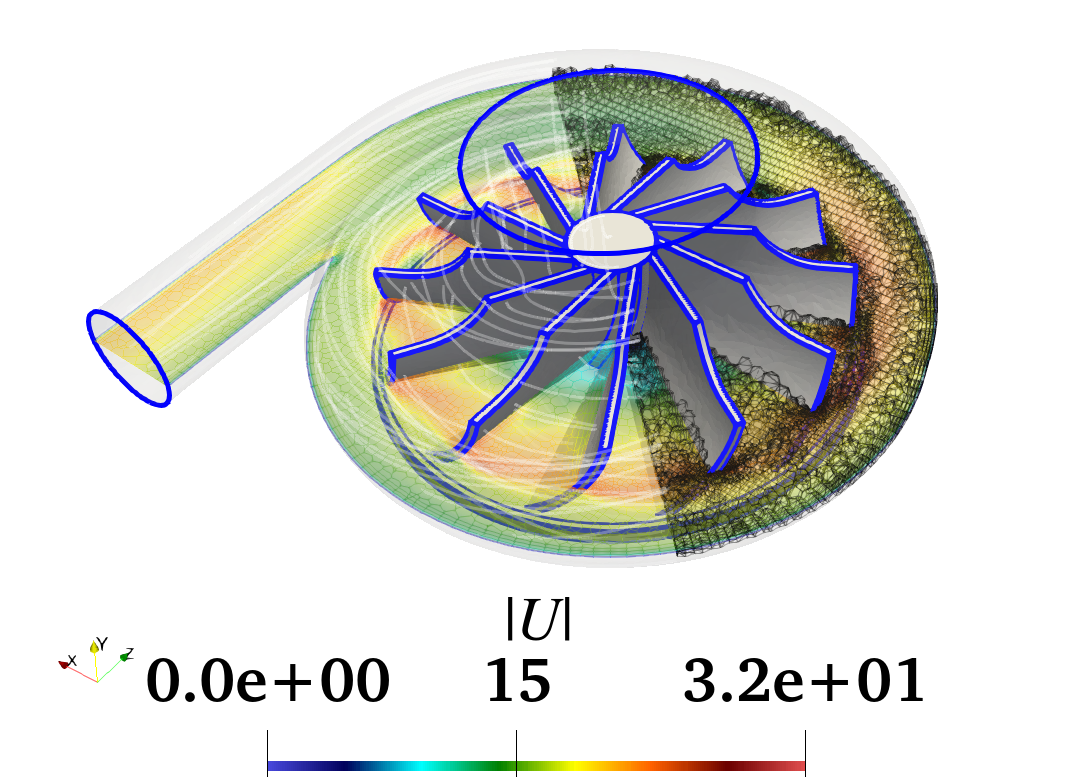}
    \caption{}\label{pump-2}
\end{subfigure}

\caption{CFD simulations created to examine model performance and
interpretability. Subfigures (a)--(c) present 2D Cylinder Obstructed
Flow, (d)--(f) present 3D Branched Pipe Flow, and (g)--(i) present 3D
Centrifugal Pump Flow, each with different design variables. Each
figure shows a velocity contour slice overlaid with the mesh and
streamlines.}
\label{datasets-figure}
\end{figure}

\section{Training Details}\label{training-appendix}

Table~\ref{tab:dataset-details} specifies the modeling task and
dataset details that are common to all models on each industrial
benchmark, Table~\ref{tab:baseline-arch} summarizes the Courant
baseline architecture, and Tables~\ref{tab:cf-hparams}--\ref{tab:pump-hparams}
report training hyperparameters and compute costs for all experiments.
All models are trained on a single NVIDIA A100 GPU on the Azure cloud
provider using NCA100v4 instances. The model architecture detailed in
\S\ref{sec:model} is implemented and trained using Pytorch
2.11~\citep{ansel2024pytorch}.

\begin{table}[htb]
  \centering
  \small
  \caption{Modeling task and dataset specification, common to all the
  models on each industrial benchmark.}
  \label{tab:dataset-details}
  \begin{tabular}{lccc}
    \toprule
                                       & Cyl.Flow & Br.Pipe & Ctr.Pump \\
    \midrule
    \multicolumn{4}{l}{\textit{Shared}} \\
    \quad Train / val / test split     & \multicolumn{3}{c}{$0.7 / 0.15 / 0.15$} \\
    \quad Random seed                  & \multicolumn{3}{c}{$42$} \\
    \quad Field normalization          & \multicolumn{3}{c}{per-component standardization} \\
    \quad Distance-field input         & \multicolumn{3}{c}{yes} \\
    \quad Input / processing domain    & \multicolumn{3}{c}{full mesh$^{\dagger}$} \\
    \midrule
    \multicolumn{4}{l}{\textit{Per dataset}} \\
    \quad Modeling task                & transient (autoregressive) & steady-state & steady-state \\
    \quad Total trajectories / samples & 200 & 300 & 300 \\
    \quad Modeled fields               & $u_x, u_y$ & $u_x, u_z$ & $u_x, u_y, u_z, p$ \\
    \quad Rollout (steps)              & $10$ & --- & --- \\
    \quad $\Delta t_{\text{pred}}$     & $10$ raw steps ($10.0$~s) & --- & --- \\
    \bottomrule
  \end{tabular}

  \medskip
  {\footnotesize $^{\dagger}$ Except Courant on the steady benchmarks
  (Br.Pipe, Ctr.Pump), which encodes only the boundary point cloud
  (Tab.~\ref{tab:baseline-arch}).}
\end{table}

\begin{table}[htb]
  \centering
  \small
  \caption{Architectural hyperparameters of the Courant baseline across
  the three industrial benchmarks. The GWA-on variant and other
  ablations are reported in Appendix~\ref{ablation-appendix}.}
  \label{tab:baseline-arch}
  \begin{tabular}{lccc}
    \toprule
                                    & Cyl.Flow & Br.Pipe & Ctr.Pump \\
    \midrule
    \multicolumn{4}{l}{\textit{Shared}} \\
    \quad Latent / hidden dim       & \multicolumn{3}{c}{$128$} \\
    \quad Attention heads (FFN mult.) & \multicolumn{3}{c}{$3$ ($\times 2$)} \\
    \quad Encoder                   & \multicolumn{3}{c}{$3$ levels, $1$ self-attention loop per level} \\
    \quad Anchor selection          & \multicolumn{3}{c}{FPS-sampled} \\
    \quad Coordinate embedding      & \multicolumn{3}{c}{RFF, shared encoder/decoder} \\
    \quad Decoder                   & \multicolumn{3}{c}{single linear cross-attention with DF-aware queries} \\
    \quad Gaussian window attention & \multicolumn{3}{c}{disabled} \\
    \quad Auxiliary regularization  & \multicolumn{3}{c}{none ($L_2$ prediction loss only)} \\
    \midrule
    \multicolumn{4}{l}{\textit{Per dataset}} \\
    \quad Number of anchors $L$     & $64$ & $256$ & $256$ \\
    \quad Latent dynamics           & NeuralODE (Euler) & --- & --- \\
    \quad Encoder input             & full mesh
                                    & boundary point cloud$^{\dagger}$
                                    & boundary point cloud$^{\dagger}$ \\
    \bottomrule
  \end{tabular}

  \medskip
  {\footnotesize $^{\dagger}$ For steady boundary-value problems the
  interior solution is determined by the boundary data, so encoding
  the boundary point cloud alone preserves predictive accuracy while
  substantially reducing the encoder's input size
  (cf.\ Table~\ref{tab:feature_ablation_nmae}, ``No boundary point
  cloud'').}
\end{table}

\begin{table}[htb]
  \centering
  \small
  \setlength{\tabcolsep}{4pt}
  \caption{Hyperparameters for the 2D experiments. All models use the
  AdamW optimizer with gradient clipping at $2.0$ and a 2-phase LR
  schedule: the initial rate is held constant over the first half of
  training, then linearly decayed to $0.1\times$ the initial LR over
  the second half. Rollout is the test-time autoregressive rollout
  length (steps of size $\Delta t_{\text{pred}}$).}
  \label{tab:cf-hparams}
  \begin{tabular}{l r r r r r r r}
    \toprule
    Model & Params & LR & Epochs & Batch & Rollout & Train (h) & GPU peak (GB) \\
    \midrule
    FNO        & 1,598,148 & 3e-4 & 300 & 100 & 10 & 4.9  & 22.2 \\
    ENF        & 1,156,546 & 3e-4 & 300 &  50 & 10 & 6.9  & 34.4 \\
    MGN        & 1,437,572 & 1e-4 & 150 &  20 & 10 & 27.8 & 42.4 \\
    UPT        & 5,158,290 & 3e-4 & 300 & 100 & 10 & 24.4 & 37.4 \\
    Transolver & 3,867,458 & 1e-3 & 100 &   8 & 10 & 11.8 & 23.1 \\
    \midrule
    Courant (baseline) & 1,298,354 & 3e-4 & 500 & 100 & 10 & 7.1 & 21.8 \\
    \bottomrule
  \end{tabular}
\end{table}

\begin{table}[htb]
  \centering
  \small
  \setlength{\tabcolsep}{4pt}
  \caption{Hyperparameters for the 3D Branched Pipe Flow
  (steady-state) experiments. Optimizer and schedule as in
  Table~\ref{tab:cf-hparams}. No autoregressive rollout is performed.}
  \label{tab:hvac-hparams}
  \begin{tabular}{l r r r r r r}
    \toprule
    Model & Params & LR & Epochs & Batch & Train (h) & GPU peak (GB) \\
    \midrule
    UPT        & 5,181,838 & 2e-4 & 2500 & 1 &  8.4 &  3.3 \\
    MGN        &   878,210 & 3e-4 &  300 & 1 &  9.4 & 78.8 \\
    Transolver & 3,865,410 & 1e-3 &  150 & 1 &  2.2 & 18.4 \\
    \midrule
    Courant (baseline) & 1,151,410 & 3e-4 & 1500 & 1 & 4.5 & 3.4 \\
    Courant (with GWA) & 1,220,286 & 3e-4 & 1500 & 1 & 4.9 & 4.2 \\
    Courant (no boundary pc) & 1,163,602 & 3e-4 & 1500 & 1 & 12.6 & 6.1 \\
    \bottomrule
  \end{tabular}
\end{table}

\begin{table}[htb]
  \centering
  \small
  \setlength{\tabcolsep}{4pt}
  \caption{Hyperparameters for the 3D Centrifugal Pump (steady-state)
  experiments. Optimizer and schedule as in
  Table~\ref{tab:cf-hparams}. No autoregressive rollout is performed.}
  \label{tab:pump-hparams}
  \begin{tabular}{l r r r r r r}
    \toprule
    Model & Params & LR & Epochs & Batch & Train (h) & GPU peak (GB) \\
    \midrule
    UPT        & 5,180,296 & 3e-4 & 300 & 1 &  4.4 & 12.6 \\
    Transolver & 3,862,852 & 1e-3 & 150 & 1 &  7.3 & 73.6 \\
    \midrule
    Courant (baseline) & 1,152,548 & 3e-4 & 300 & 1 & 2.3 & 14.6 \\
    Courant (with GWA) & 1,221,424 & 3e-4 & 300 & 1 & 2.4 & 18.9 \\
    Courant (no boundary pc) & 1,156,612 & 3e-4 & 300 & 1 & 8.3 & 26.4 \\
    \bottomrule
  \end{tabular}
\end{table}

\section{Ablation Studies}\label{ablation-appendix}

\begin{table}[htb]
\caption{Feature ablation (NMAE). The baseline includes: FPS-sampled
latent anchors, distance field as a decoder query feature, shared
coordinate embedding, scalar global features (industrial only), and
boundary point cloud (industrial only).}
\label{tab:feature_ablation_nmae}
\centering
\begin{tabular}{@{}lccc@{}}
\toprule
& \multicolumn{2}{c}{Industrial RANS} & Unsteady \\
\cmidrule(lr){2-3}\cmidrule(l){4-4}
& Br.Pipe & Ctr.Pump & Cyl.Flow \\
\midrule
Baseline                 & \sn{9.7}{2}          & \sn{1.8}{1}          & \first{\sn{1.7}{1}} \\
GWA (enc + dec)          & \sn{1.1}{1}          & \second{\sn{1.7}{1}} & --- \\
GWA in decoder only      & \sn{1.0}{1}          & \second{\sn{1.7}{1}} & --- \\
GWA in encoder only      & \sn{1.0}{1}          & \sn{1.8}{1}          & --- \\
Learned anchors          & \first{\sn{8.9}{2}}  & \first{\sn{1.6}{1}}  & \sn{1.9}{1} \\
Abstract learned queries & \second{\sn{9.4}{2}} & \first{\sn{1.6}{1}}  & \sn{1.9}{1} \\
No shared coord emb.     & \sn{9.6}{2}          & \sn{1.8}{1}          & \sn{1.9}{1} \\
No distance field        & \sn{1.9}{1}          & \sn{2.2}{1}          & \sn{2.1}{1} \\
No global data           & \sn{1.0}{1}          & \sn{1.8}{1}          & --- \\
No boundary PC (full-mesh) & \sn{9.6}{2}        & \sn{1.9}{1}          & --- \\
MLP ODE integrator       & ---                  & ---                  & \second{\sn{1.8}{1}} \\
\bottomrule
\end{tabular}

\medskip
{\footnotesize --- = ablation not applicable to dataset.}
\end{table}

Table~\ref{tab:feature_ablation_nmae} reports a per-component ablation
of the baseline configuration of Tab.~\ref{tab:baseline-arch}. Dashes
mark ablations that do not apply to a given dataset.

\paragraph{Architectural choices}\label{sec:arch-choices}
Removing the distance-field aware decoder query gives the largest
single-component degradation on every dataset, increasing NMAE by
$30$--$40\%$. Between the two geometry-anchored variants ---
\emph{FPS-sampled} and \emph{learnable} anchors (both with
coordinates embedded through the shared RFF) --- FPS-sampled anchors
are best for Cyl.Flow; on Br.Pipe and Ctr.Pump, learnable anchors
edge out FPS. Removing the shared encoder/decoder coordinate embedding
has its largest effect on Cyl.Flow
($1.7\!\to\!1.9\!\times\!10^{-1}$) and a small effect on the steady
cases. Removing the global-parameter conditioning costs a small amount
on Br.Pipe ($9.7\!\to\!1.0\!\times\!10^{-1}$) and is neutral on
Ctr.Pump. Replacing the self-attention NeuralODE integrator with an
$\MLP$ on Cyl.Flow costs accuracy
($1.7\!\to\!1.8\!\times\!10^{-1}$). Adding Gaussian-window attention
to the steady benchmarks slightly hurts Br.Pipe
($9.7\!\to\!1.0\!\times\!10^{-1}$) and gives a small decoder-side
gain on Ctr.Pump ($1.8\!\to\!1.7\!\times\!10^{-1}$); it visibly
tightens the spatial support (Fig.~\ref{decomposition-cf-figure}).

\paragraph{Boundary Point Cloud (PC) as encoder input}\label{sec:encoder-bpc}
Encoding only the boundary point cloud preserves accuracy on both
steady benchmarks (within $5\%$ of baseline NMAE) while reducing
training time from $12.6$~h to $4.5$~h and GPU peak memory from
$6.1$~GB to $3.4$~GB on Br.Pipe (Tab.~\ref{tab:hvac-hparams}), and
from $8.3$~h to $2.3$~h and $26.4$~GB to $14.6$~GB on Ctr.Pump
(Tab.~\ref{tab:pump-hparams}).

\paragraph{Memory and compute efficiency}\label{sec:resource-efficiency}
Across all three benchmarks
(Tabs.~\ref{tab:cf-hparams}--\ref{tab:pump-hparams}) Courant's
parameter count ($\sim\!1.2\!-\!1.3$M) is smaller than Transolver
and UPT ($3\!-\!5\times$). Its GPU peak memory is lower than UPT
(e.g., $21.8$~GB vs $37.4$~GB on Cyl.Flow) and MGN ($78.8$~GB on Br.Pipe, $42.4$~GB on Cyl.Flow) and
Transolver ($73.6$~GB on Ctr.Pump). Training time follows the same
pattern, with Courant $1.5$--$3\times$ faster than the heaviest
transformer baselines: $7.1$~h vs $24.4$~h for UPT on Cyl.Flow, and
$2.3$~h vs $4.4$~h for UPT and $7.3$~h for Transolver on Ctr.Pump.

\begin{figure}[htb]
    \centering
    \includegraphics[width=1\linewidth]{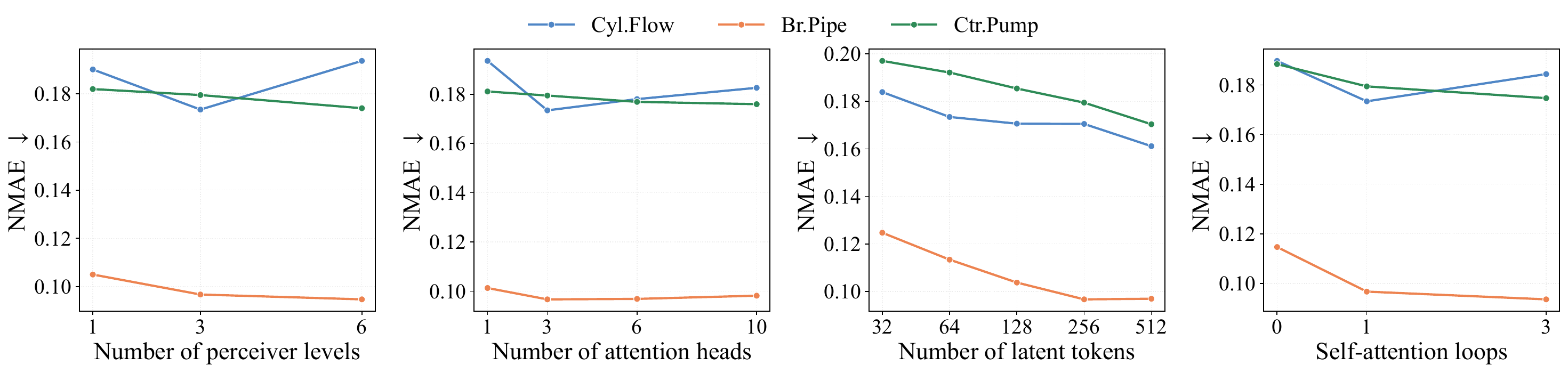}
    \caption{Ablation study sweeping each architectural hyperparameter
    in isolation while holding the others at the baseline values of
    Tab.~\ref{tab:baseline-arch}. The number of latent tokens $L$ has
    the largest and most consistent effect: NMAE decreases
    monotonically with $L$ on all three datasets, saturating near
    $L\!=\!256$ on Br.Pipe. Perceiver levels and attention heads
    saturate at three; the mild U-shape on Cyl.Flow is consistent with
    over-parameterization on its smaller mesh. Self-attention loops
    bring most of their benefit on the $0\!\to\!1$ jump, with
    diminishing returns thereafter. The baseline configuration sits at
    or near the per-dataset optimum in every panel.}
    \label{fig:ablations}
\end{figure}

\section{Latent Basis Decomposition: Single-Head Reformulation and POD Analogy}\label{decomposition-appendix}

We derive here the single-head, no-$\LayerNorm$ version of the
per-token decoder decomposition discussed in \S\ref{sec:decoder}, and
contrast it with the classical proper-orthogonal-decomposition (POD)
basis projection. Let $\mathbf{Z}_t = \{\mathbf{z}_l(t)\}_{l=1}^{L}$
denote the latent tokens at time $t$, and let
$\{\mathbf{x}_n\}_{n=1}^{N}$ denote the query coordinates. For a
single-head linear cross-attention decoder, define
\begin{equation}
\mathbf{q}_n = W_Q\,\gamma(\mathbf{x}_n),\qquad
\mathbf{k}_l(t) = W_K\,\mathbf{z}_l(t),\qquad
\alpha_{nl}(t) = \mathrm{softmax}_l\!\Bigl(
  \tfrac{\mathbf{q}_n^{\!\top}\mathbf{k}_l(t)}{\sqrt{d_k}}\Bigr).
\label{eq:single-head-attn}
\end{equation}
The decoder output is then
\begin{equation}
\hat{\mathbf{u}}(\mathbf{x}_n,t)
= \sum_{l=1}^{L} \alpha_{nl}(t)\,\mathbf{b}_l(t),
\qquad \mathbf{b}_l(t) = W_V\,\mathbf{z}_l(t),
\label{eq:single-head-out}
\end{equation}
which can be rewritten as a separable expansion
\begin{equation}
\hat{\mathbf{u}}(\mathbf{x},t)
= \sum_{l=1}^{L} \mathbf{b}_l(t)\,\phi_l(\mathbf{x},t),
\qquad \phi_l(\mathbf{x},t) = \alpha_l(\mathbf{x},t).
\label{eq:single-head-basis}
\end{equation}
This resembles the classical POD expansion
\begin{equation}
u(\mathbf{x},t) \;\approx\; \sum_{r=1}^{R} a_r(t)\,\varphi_r(\mathbf{x}),
\label{eq:pod-expansion}
\end{equation}
with one important distinction. In POD, the spatial basis functions
$\varphi_r(\mathbf{x})$ are fixed across time, learned from a
snapshot matrix once and reused. In the cross-attention decoder, both
the coefficients $\mathbf{b}_l(t)$ \emph{and} the spatial functions
$\phi_l(\mathbf{x},t)$ depend on the evolving latent state, so the
basis itself is state-dependent.

\section{Additional experimental details}\label{experiment_details-appendix}

\subsection{Additional benchmarks}\label{community-benchmark-appendix}

We evaluate on five additional benchmarks from the Neural Solver
Library~\citep{wu2024transolver} spanning steady and time-dependent
regimes on structured and unstructured meshes.
Table~\ref{tab:community-benchmarks-overview} summarizes the benchmark
suite.

\begin{table}[htb]
\caption{Benchmark overview.}
\label{tab:community-benchmarks-overview}
\centering
\small
\begin{tabular}{@{}llllll@{}}
\toprule
Dataset & Regime & Geometry & Features & Targets & Prediction \\
\midrule
Elasticity  & Steady   & Unstructured 2D & 0 (+2 coords) & 1 & Single-step \\
Pipe flow   & Steady   & Structured 2D   & 2             & 1 & Single-step \\
Airfoil     & Steady   & Structured 2D   & 2             & 1 & Single-step \\
Darcy flow  & Steady   & Structured 2D   & 1             & 1 & Single-step \\
Diff-Sorp (PDEBench) & Time-dep. & Structured 1D & 10 & 1 & Latent ODE \\
3D CFD (PDEBench)    & Time-dep. & Structured 3D & 50 & 5 & Latent ODE \\
\bottomrule
\end{tabular}

\medskip
{\footnotesize For Elasticity, Courant appends the 2D spatial
coordinates to the input features; baselines use the original
zero-dimensional input. Time-dependent datasets provide 10 timesteps
of input with N targets each, hence 50 input features and 5 targets
noted for 3D CFD. The benchmark predicts the next 10 timesteps.}
\end{table}

\paragraph{Courant architecture on additional benchmarks}
The architecture follows the same encoder--processor--decoder design
described in \S\ref{sec:encoder}--\ref{sec:decoder}. A single
configuration is used across all benchmarks
(Table~\ref{tab:courant-shared-hparams}), with the only per-dataset
variation being the number of latent tokens: 256 for steady-state
problems and 64 for the two time-dependent benchmarks. For the
time-dependent benchmarks (Diff-Sorp, 3D CFD), the encoder is applied
once to $T_{\mathrm{in}}=10$ input frames to produce an initial
latent state, and the NeuralODE integrator (\S\ref{sec:processor})
advances this state forward $T_{\mathrm{out}}=10$ steps with Euler
integration at $\Delta\tau = 1.0$. The decoder is applied at each
step to produce field predictions.

\begin{table}[htb]
\caption{Courant architectural hyperparameters on community
benchmarks. A single configuration is used throughout; the only
variation is the number of latent tokens (256 for steady, 64 for
time-dependent).}
\label{tab:courant-shared-hparams}
\centering
\small
\begin{tabular}{@{}lr@{}}
\toprule
Hyperparameter & Value \\
\midrule
Hidden / latent dimension  & 128 \\
Attention heads            & 3 \\
Encoder levels             & 6 \\
FF multiplier              & 2 \\
Latent query initialisation & Furthest point sampling \\
Coordinate embedding       & RFF ($\sigma=0.3$, learnable) \\
Number of latents (steady) & 256 \\
Number of latents (time-dependent) & 64 \\
\bottomrule
\end{tabular}
\end{table}

\paragraph{Training configuration}
All experiments use AdamW with cosine annealing (except Diff-Sorp,
which uses StepLR with step\,=\,100 and $\gamma$\,=\,0.5). The loss
is the relative $L_2$ error summed over the batch and normalised per
sample by the ground-truth norm. For Darcy flow, a derivative
regularisation term (weight 0.1) computed via central finite
differences is added. Table~\ref{tab:courant-training} reports the
full per-dataset training configuration. Courant and UPT are trained
for 1000 epochs on the steady-state benchmarks, while other baselines
use 500 epochs following the Neural Solver Library defaults. All
experiments run on a single NVIDIA A100 GPU on the Azure cloud
provider using NCA100v4 instances. The maximum required training time
is $\sim$30 hours, with many benchmarks requiring fewer than 6 hours.

\begin{table}[htb]
\caption{Per-dataset training configuration for Courant on community
benchmarks. Default weight decay is $10^{-5}$ and default scheduler
is CosineAnnealingLR unless noted.}
\label{tab:courant-training}
\centering
\small
\begin{tabular}{@{}lrrrrll@{}}
\toprule
Dataset & Epochs & Batch & LR & Weight decay & Scheduler & Train / Test \\
\midrule
Elasticity  & 1000 &  1 & 5e-5 & 1e-5 & Cosine & 1000 / 200 \\
Pipe        & 1000 &  4 & 1e-4 & 1e-5 & Cosine & 1000 / 200 \\
Airfoil     & 1000 &  8 & 5e-4 & 1e-4 & Cosine & 1000 / 200 \\
Darcy       & 1000 &  4 & 5e-4 & 1e-5 & Cosine & 1000 / 200 \\
Diff-Sorp   &  500 & 20 & 5e-4 & 1e-4 & StepLR & 9000 / 1000 \\
3D CFD      &  500 &  2 & 5e-4 & 1e-4 & Cosine & 1000 / 200 \\
\bottomrule
\end{tabular}

\medskip
{\footnotesize Airfoil uses gradient clipping (max norm\,=\,0.1).
Darcy uses derivative loss (weight 0.1) and $5{\times}$ spatial
downsampling per dimension ($85{\times}85$ grid). 3D CFD uses
$8{\times}$ spatial downsampling in all three dimensions. Diff-Sorp
uses the standard PDEBench split (9000 / 1000).}
\end{table}

\paragraph{Baseline configurations}
All baselines use the original Neural Solver Library hyperparameters
with the same loss function (relative $L_2$, plus derivative loss
where applicable). Default baseline settings are: AdamW with
LR\,=\,$10^{-3}$, weight decay\,=\,$10^{-5}$, batch size\,=\,8, 500
epochs, and CosineAnnealingLR with a 1000/200 train/test split. UPT
deviates from these defaults by training for 1000 epochs.
Tables~\ref{tab:baseline-transolver}--\ref{tab:baseline-factformer}
report per-model configurations.

\begin{table}[htb]
\caption{Transolver baseline configuration. All datasets use hidden
dim\,=\,128, 8 heads, 8 layers, and 64 slices.}
\label{tab:baseline-transolver}
\centering
\small
\begin{tabular}{@{}lrrrlll@{}}
\toprule
Dataset & Epochs & Batch & LR & Weight decay & Scheduler & Notes \\
\midrule
Elasticity  & 500 &  1 & 1e-3 & 1e-5 & Cosine & --- \\
Pipe        & 500 &  4 & 1e-3 & 1e-5 & Cosine & --- \\
Airfoil     & 500 &  8 & 1e-3 & 1e-4 & Cosine & Grad clip 0.1 \\
Darcy       & 500 &  4 & 1e-3 & 1e-5 & Cosine & Derivative loss \\
Diff-Sorp   & 500 & 20 & 5e-4 & 1e-4 & StepLR & AR rollout \\
\bottomrule
\end{tabular}
\end{table}

\begin{table}[htb]
\caption{FNO baseline configuration.}
\label{tab:baseline-fno}
\centering
\small
\begin{tabular}{@{}lrrrlll@{}}
\toprule
Dataset & Hidden & Layers & Epochs & Batch & LR & Notes \\
\midrule
Elasticity  & 32  & 8 & 500 &  4 & 1e-3 & --- \\
Pipe        & 128 & 8 & 500 &  4 & 1e-3 & --- \\
Airfoil     & 32  & 8 & 500 &  4 & 1e-3 & --- \\
Darcy       & 128 & 8 & 500 &  4 & 1e-3 & Derivative loss \\
Diff-Sorp   & 64  & 8 & 500 & 20 & 5e-4 & AR rollout \\
3D CFD      & 20  & 8 & 500 &  5 & 5e-4 & AR rollout, $8{\times}$ downsample \\
\bottomrule
\end{tabular}
\end{table}

\begin{table}[htb]
\caption{UPT baseline configuration. All datasets use hidden
dim\,=\,512, 8 heads, 6 layers, and 32 output tokens. UPT is trained
for 1000 epochs on all benchmarks.}
\label{tab:baseline-upt}
\centering
\small
\begin{tabular}{@{}lrrrll@{}}
\toprule
Dataset & Epochs & Batch & LR & Weight decay & Notes \\
\midrule
Elasticity & 1000 & 1 & 5e-4 & 1e-5 & --- \\
Pipe       & 1000 & 4 & 5e-4 & 1e-5 & --- \\
Airfoil    & 1000 & 8 & 5e-4 & 1e-5 & --- \\
Darcy      & 1000 & 4 & 5e-4 & 1e-5 & Derivative loss \\
Diff-Sorp  & 1000 & 20 & 5e-4 & 1e-5 & AR rollout, StepLR \\
Darcy      & 1000 & 4 & 5e-4 & 1e-5 & Derivative loss \\
Diff-Sorp  & 1000 & 16 & 5e-4 & 1e-4 & AR rollout, StepLR \\
\bottomrule
\end{tabular}
\end{table}

\begin{table}[ht]
\caption{FactFormer baseline configuration. All use 8 heads, 8
layers, 500 epochs, batch 4, LR\,=\,$10^{-3}$.}
\label{tab:baseline-factformer}
\centering
\small
\begin{tabular}{@{}lrrrl@{}}
\toprule
Dataset & Hidden & Batch & LR & Notes \\
\midrule
Airfoil & 32  & 4 & 1e-3 & --- \\
Pipe    & 128 & 4 & 1e-3 & --- \\
Darcy   & 128 & 4 & 1e-3 & Derivative loss \\
\bottomrule
\end{tabular}
\end{table}

\paragraph{Evaluation protocol}
All models are evaluated using the relative $L_2$ error averaged over
the test set. For temporal benchmarks, Courant uses latent ODE rollout
over $T_{\mathrm{out}}=10$ steps from $T_{\mathrm{in}}=10$ input
frames, while baselines use autoregressive rollout in observation
space; error is reported over all predicted steps jointly. Full
results are reported in Tables~\ref{tab:bench-unstr} and
\ref{tab:bench-str}.

\section{Frequency-domain analysis of the wake and the latent trajectory}\label{vortex-shedding-section}

\begin{figure}[htb]
    \centering
    \includegraphics[width=1\linewidth]{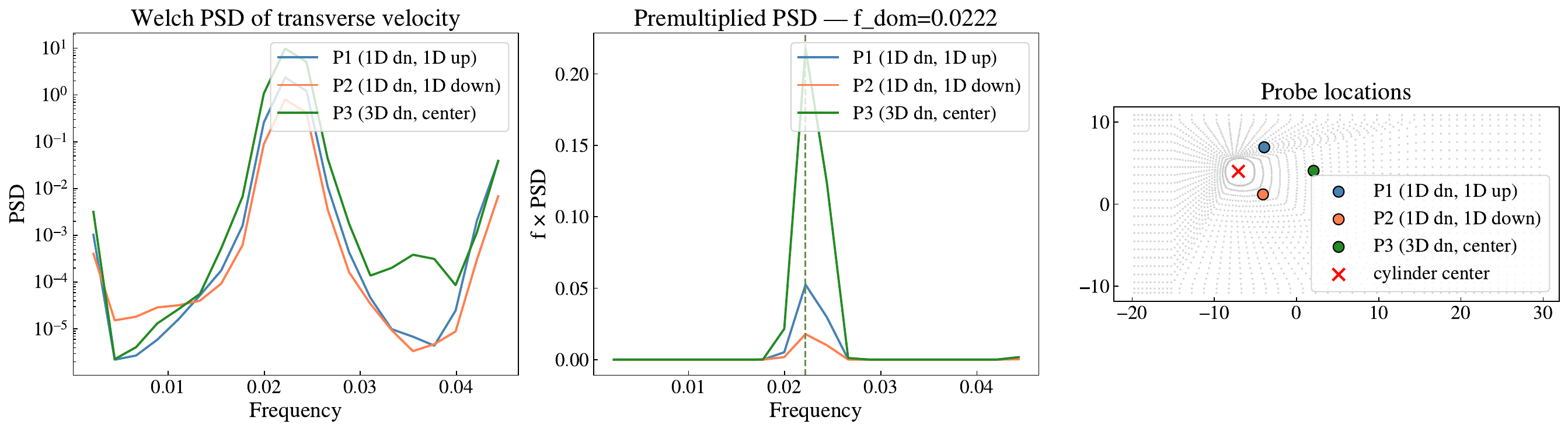}
    \caption{Wake-probe vortex shedding frequency. Welch PSD of the
    transverse velocity at three downstream probes (left),
    premultiplied $f\!\cdot\!\mathrm{PSD}$ identifying the dominant
    peak at $f_{\text{shed}}\!=\!0.0222$ (centre), and probe
    locations relative to the cylinder (right). All three probes
    collapse onto the same peak, consistent with laminar K\'arm\'an
    shedding.}
    \label{fig:cf_shedding_frequency}
\end{figure}

Both the main-text latent PSD (Fig.~\ref{fig:cf_latent_psd}) and the
wake-probe reference (Fig.~\ref{fig:cf_shedding_frequency}) are
estimated using Welch's modified periodogram method with a Hann window
and 50\% overlap; we summarize the two pipelines below for
reproducibility.

\paragraph{Wake-probe PSD}
We instrument three Eulerian probes in the cylinder wake at offsets
$(\Delta x,\Delta y)\!=\!(1D,\!+\!1D)$, $(1D,\!-\!1D)$, and
$(3D,0)$ relative to the cylinder centre, and record the transverse
velocity component over the simulation horizon. Each probe series is
mean-subtracted, and a PSD is estimated with segment length
$N_{\text{seg}}\!=\!\min(N/2,\,64)$ where $N$ is the number of
snapshots. The dominant frequency is read off as the $\arg\max$ of
the positive-frequency PSD (excluding $f\!=\!0$) and averaged across
probes. All three probes report the same peak at
$f_{\text{shed}}\!\approx\!0.0222$
(Fig.~\ref{fig:cf_shedding_frequency}, centre panel), consistent with
the laminar K\'arm\'an shedding regime expected at
$\mathrm{Re}\!\approx\!90$.

\paragraph{Latent PSD}
For Fig.~\ref{fig:cf_latent_psd} we run a single ODE rollout of
$T\!=\!40$ steps from one starting slice; because latents are
advanced in place by the integrator, they remain mutually aligned
across time. We fit a $k\!=\!3$ PCA on the latent feature dimension
using only the in-training-horizon segment of the trajectory (the
first ten steps) as the basis, then project every latent anchor onto
the leading components for the full rollout. Welch's method is
applied per latent anchor--PC pair using the same physical sampling
rate $f_s\!=\!1/\Delta t_{s}$. To keep Fig.~\ref{fig:cf_latent_psd}
legible we display the seven latent anchors with the highest PSD at
$f_{\text{shed}}$ on PC0 together with the three lowest, illustrating
the contrast between latents that carry the shedding signal and those
that do not.

\paragraph{Latent scalogram and drift diagnostic}
Figure~\ref{fig:cf_latent_scalogram} resolves the same ODE-evolved
latent trajectory in time and frequency simultaneously. We apply a
complex Morlet continuous wavelet transform with $\omega_0\!=\!6$
cycles per wavelet, evaluated on a $64$-bin log-spaced frequency grid
bounded below by $\max(f_s/T,\,f_{\text{shed}}/8)$ and above by
$\min(0.45\,f_s,\,4\,f_{\text{shed}})$, so that the bracket spans
roughly an octave below to two octaves above the shedding peak. The
transform is FFT-based and analytic (negative frequencies zeroed),
with input mean-subtracted and zero-padded to the next power of two.
Magnitudes are rendered in decibels. At each time bin we read off the
peak frequency
$f_{\text{peak}}(t)\!=\!\arg\max_f |W(t,f)|$ and report the relative
drift $(f_{\text{peak}}(t)-f_{\text{shed}})/f_{\text{shed}}$ together
with a $\pm 20\%$ visual reference band.

\subsection{NeuralODE Jacobian eigenspectrum}\label{sec:eigenspectrum-appendix}

\begin{figure}[htb]
    \centering
    \includegraphics[width=0.5\linewidth]{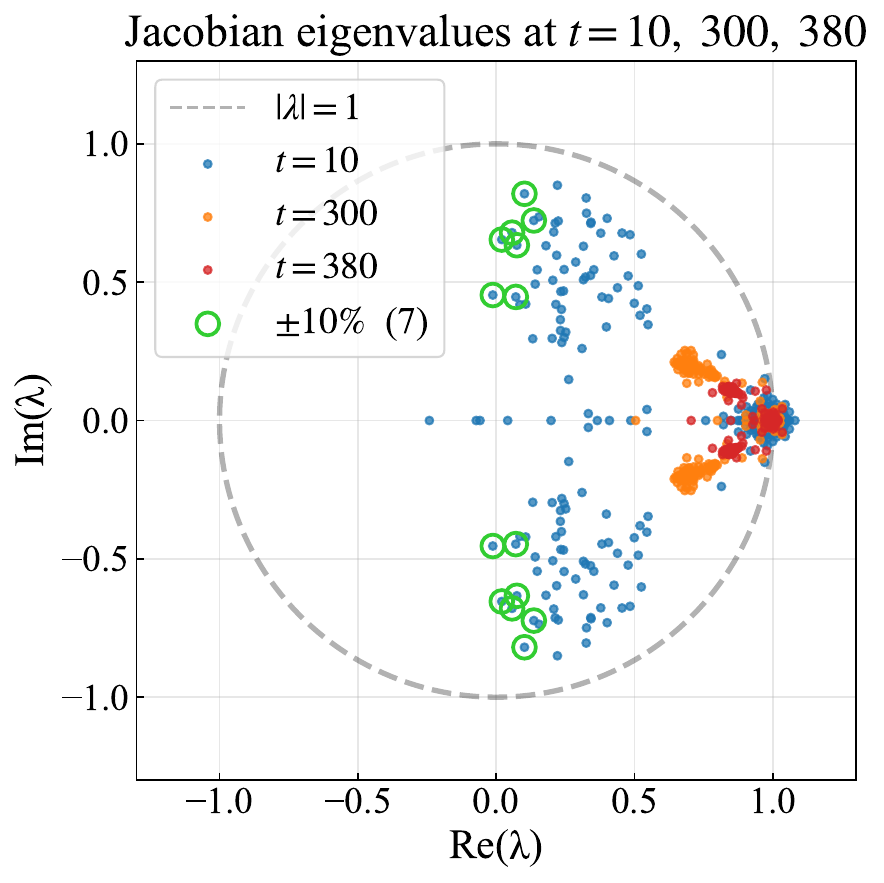}
    \caption{Eigenvalues of the NeuralODE Jacobian
    $\partial f_\theta / \partial \mathbf{Z}$ at integration times
    $t\!=\!10,300,380$, plotted in the complex plane. At $t\!=\!10$ a
    complex-conjugate eigenpair near $|\lambda|\!\approx\!1$ at
    $\mathrm{Im}(\lambda)\!\approx\!\pm 0.65$ (highlighted,
    $\pm 10\%$ rings) corresponds to the shedding-frequency
    oscillatory mode. By $t\!=\!300,380$ the spectrum collapses
    toward the real axis as the rollout relaxes into a stationary
    regime.}
    \label{fig:cf_jacobian_eigenvalues}
\end{figure}

The dynamical content of the rollout is also exposed in the
operator's spectrum. Figure~\ref{fig:cf_jacobian_eigenvalues} plots
the eigenvalues of the NeuralODE Jacobian
$\partial f_\theta / \partial \mathbf{Z}$ at three integration times
in the complex plane. At $t\!=\!10$ the spectrum contains a
complex-conjugate eigenpair near $|\lambda|\!\approx\!1$ at
$\mathrm{Im}(\lambda)\!\approx\!\pm 0.65$ (highlighted,
$\pm 10\%$ rings) --- the operator-level signature of an oscillatory
mode whose frequency $\omega \!=\! \arg(\lambda)/\Delta\tau$ matches
the shedding frequency. By $t\!=\!300$ and $t\!=\!380$ the
eigenvalues collapse toward the real axis, consistent with the latent
rollout relaxing into a stationary regime.

Together with the PSD and scalogram analyses in
\S\ref{sec:resultsanddiscussion}, this shows that the shedding
frequency is exposed both in the latent trajectory and in the
operator's spectrum, making physics-grounded diagnostics directly
available for rollout-quality monitoring without retraining or
auxiliary modules.

\section{Gaussian window adaptive attention bias}%
\label{sec:gwa-appendix}%
\label{sec:gwa}

Anchors are updated across layers through a zero-initialised residual
driven by the current latent,
\begin{equation}
\mathbf{p}_j^{(\ell)} = \mathbf{p}_j^{(0)} + \Delta_\ell\!\bigl(\mathbf{z}_j^{(\ell)}\bigr),
\label{eq:anchor-update}
\end{equation}
where $\Delta_\ell$ is a small $\MLP$ with zero-initialised last
layer. Training therefore starts from a learned initial anchor cloud
of Eq.~\eqref{eq:anchor-init} and anchors are steered toward
geometrically salient regions (wakes, stagnation points, shocks,
boundary layers). Combined with the learnable $\sigma(\ell,j)$, this
in principle allows an anchor to track a moving feature.

The $\CrossAttn^{(\ell)}$ above is modulated by Gaussian window
adaptive attention (GWA), which uses $\sigma(\ell,j)$ to adapt the
center of a Gaussian window.

We add a learnable, spatially localized adaptive bias to the
pre-softmax logits: for a latent with anchor $\mathbf{p}_j^{(\ell)}$
and a key at coordinate $\mathbf{x}_i$,
\begin{equation}
b_{j,i}^{(\ell)} = -\,
\frac{\bigl\|\mathbf{p}_j^{(\ell)}-\mathbf{x}_i\bigr\|^{2}}
     {\sigma^{2}(\ell,j)},
\qquad
\sigma(\ell,j) = \sigma_0\,\kappa(\ell,j),\quad
\sigma_0 = \alpha\,\tfrac{\mathrm{domain}(\mathbf{x})}{\sqrt{L}},
\label{eq:gwa}
\end{equation}
which is added to the scaled-dot-product logit in every encoder
$\CrossAttn^{(\ell)}$ and in the decoder cross-attention (with
$\mathbf{x}_i$ replaced by the decoder query coordinate). Here
$\sigma_0$ is a dataset-scale initialization that approximately tiles
the domain with $L$ overlapping windows, and $\kappa(\ell,j)$ is a
learnable multiplier.

\newpage
\section{Additional visualizations}\label{appendix-decompositions}

In this appendix we provide additional reference visualization of
model predictions and decomposed fields. For each dataset we show two
representative test samples, showing predicted vs ground truth field
and the top 8 norm-field components of the decomposed field. For
steady-state datasets we also show examples with GWA modulation.

\clearpage
\subsection{2D Cylinder Obstructed Flow}\label{sec:2d-cylinder-obstructed-flow-visualizations}

\begin{figure}[htb]
\centering
\begin{subfigure}[b]{0.48\textwidth}
    \centering
    \includegraphics[width=\linewidth]{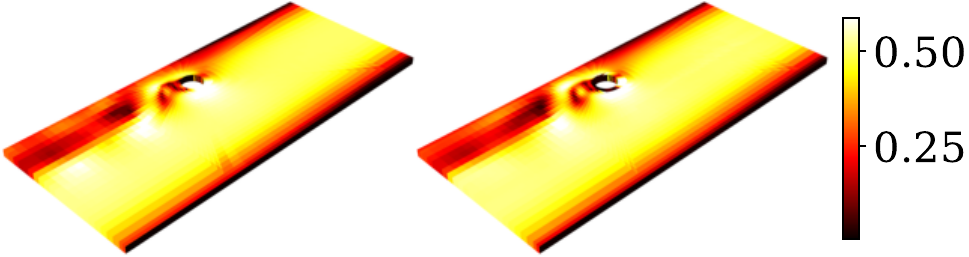}
    \caption{S1: truth vs.\ pred.}
    \label{fig:cf-s1-truepred}
\end{subfigure}\hfill
\begin{subfigure}[b]{0.48\textwidth}
    \centering
    \includegraphics[width=\linewidth]{figures/appendix/cf_per_latent_decode_sample2_true_pred_head_basis.pdf}
    \caption{S2: truth vs.\ pred.}
    \label{fig:cf-s2-truepred}
\end{subfigure}
\begin{subfigure}[b]{1.0\textwidth}
    \centering
    \includegraphics[width=\linewidth]{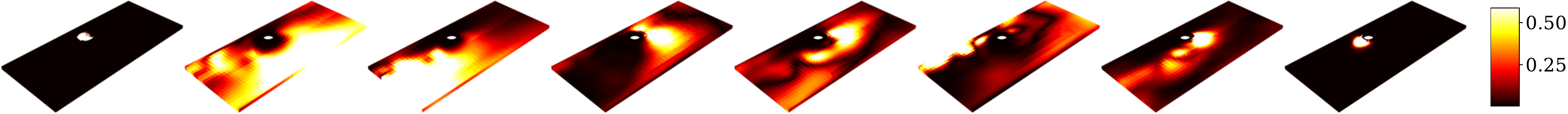}
    \caption{S1: top-$\|\boldsymbol{\delta}_k\|$ decomposition.}
    \label{fig:cf-s1-norm}
\end{subfigure}
\begin{subfigure}[b]{1.0\textwidth}
    \centering
    \includegraphics[width=\linewidth]{figures/appendix/cf_per_latent_decode_sample2_norm_head_basis.pdf}
    \caption{S2: top-$\|\boldsymbol{\delta}_k\|$ decomposition.}
    \label{fig:cf-s2-norm}
\end{subfigure}
\begin{subfigure}[b]{0.48\textwidth}
    \centering
    \includegraphics[width=\linewidth]{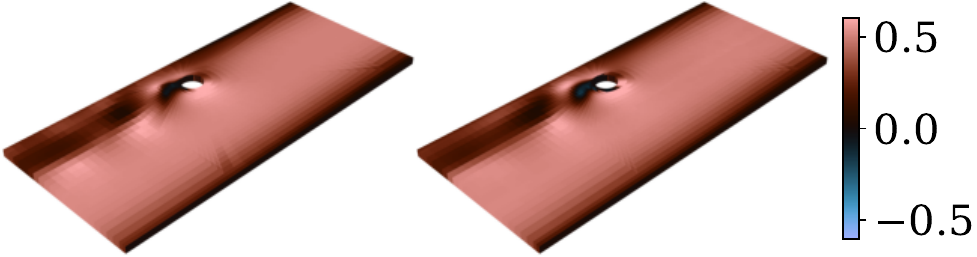}
    \caption{S1: predicted $u_x$.}
    \label{fig:cf-s1-vel-x}
\end{subfigure}\hfill
\begin{subfigure}[b]{0.48\textwidth}
    \centering
    \includegraphics[width=\linewidth]{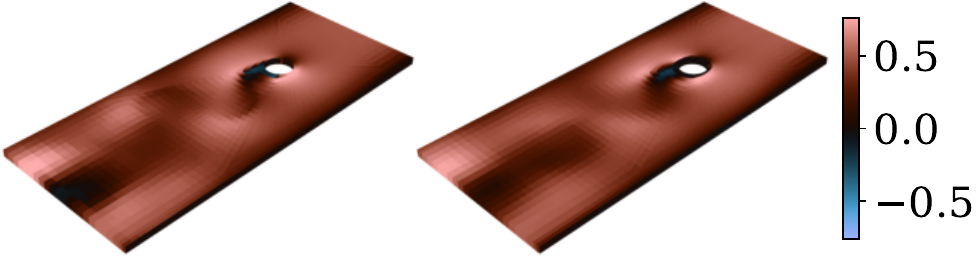}
    \caption{S2: predicted $u_x$.}
    \label{fig:cf-s2-vel-x}
\end{subfigure}\hfill
\begin{subfigure}[b]{0.48\textwidth}
    \centering
    \includegraphics[width=\linewidth]{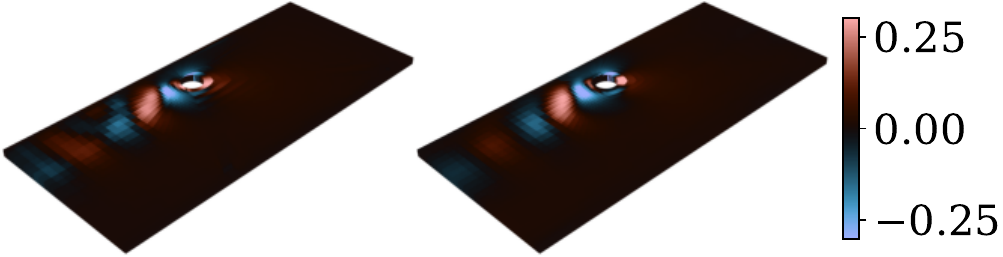}
    \caption{S1: predicted $u_y$.}
    \label{fig:cf-s1-vel-y}
\end{subfigure}\hfill
\begin{subfigure}[b]{0.48\textwidth}
    \centering
    \includegraphics[width=\linewidth]{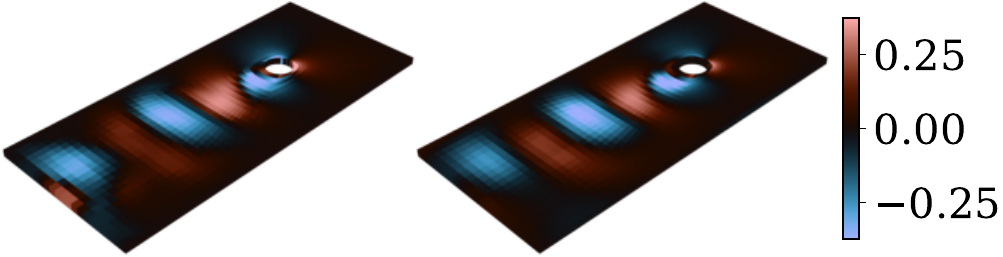}
    \caption{S2: predicted $u_y$.}
    \label{fig:cf-s2-vel-y}
\end{subfigure}
\caption{Latent basis decomposition for the 2D Cylinder-Obstructed
Flow at final time step t=10.
(a,\,b)~Ground truth and predicted velocity for two test samples.
(c,\,d)~Top-norm per-anchor decoded fields $\boldsymbol{\delta}_k$,
ranked left to right. The decomposed contributions partition the
domain into compact regions aligned with the cylinder boundary layer
and wake structure.
(e--h)~Full predicted velocity field by component ($u_x$, $u_y$)
for each sample.}
\label{fig:cf-flow-modes}
\end{figure}

\clearpage
\subsection{3D Centrifugal Pump}\label{sec:3d-centrifugal-pump-visualizations}

\begin{figure}[h]
\centering
\begin{subfigure}[b]{0.4\textwidth}
    \centering
    \includegraphics[width=\linewidth]{figures/appendix/pump_per_latent_decode_sample1_true_pred_head_basis.pdf}
    \caption{S1: truth vs.\ pred.}
    \label{fig:pump-s1-truepred}
\end{subfigure}
\begin{subfigure}[b]{0.4\textwidth}
    \centering
    \includegraphics[width=\linewidth]{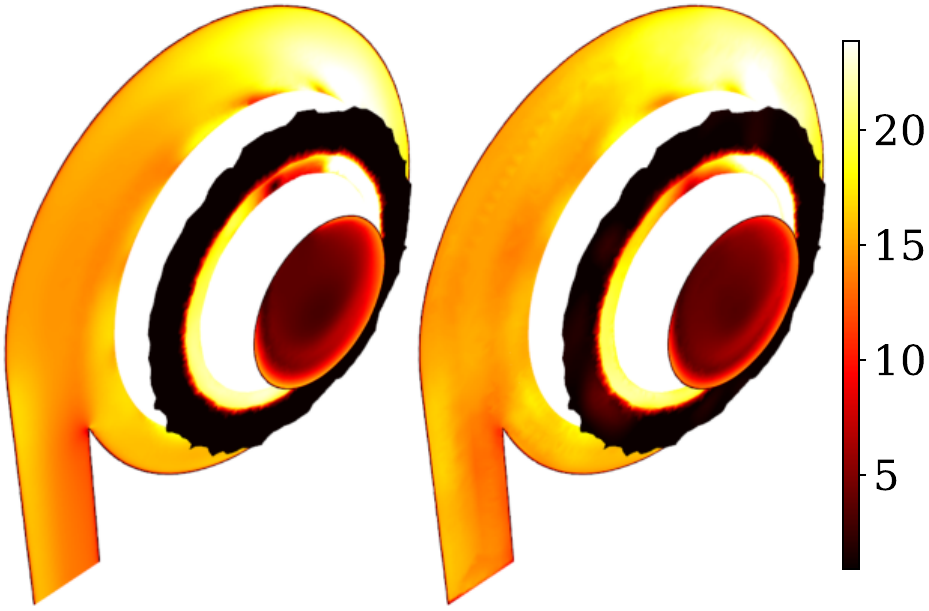}
    \caption{S2: truth vs.\ pred.}
    \label{fig:pump-s2-truepred}
\end{subfigure}
\begin{subfigure}[b]{1.0\textwidth}
    \centering
    \includegraphics[width=\linewidth]{figures/appendix/pump_per_latent_decode_sample1_norm_head_basis.pdf}
    \caption{S1: top-$\|\boldsymbol{\delta}_k\|$ decomposition.}
    \label{fig:pump-s1-norm}
\end{subfigure}
\begin{subfigure}[b]{1.0\textwidth}
    \centering
    \includegraphics[width=\linewidth]{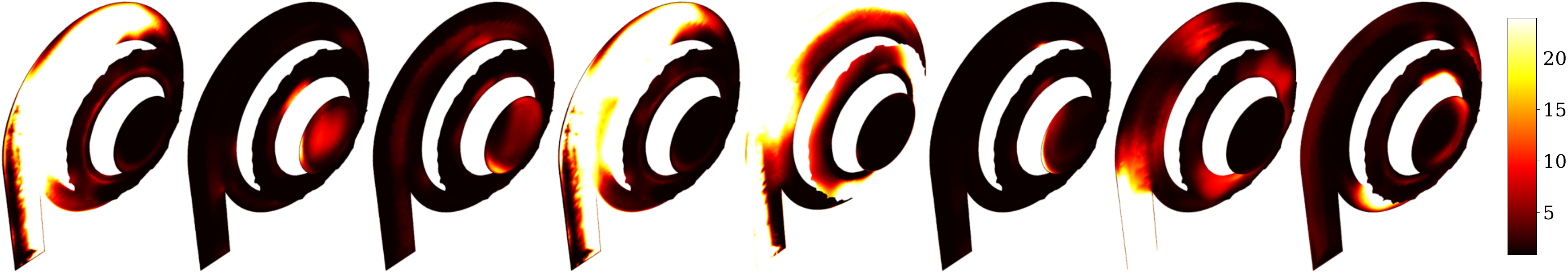}
    \caption{S2: top-$\|\boldsymbol{\delta}_k\|$ decomposition.}
    \label{fig:pump-s2-norm}
\end{subfigure}
\begin{subfigure}[b]{0.2\textwidth}
    \centering
    \includegraphics[width=\linewidth]{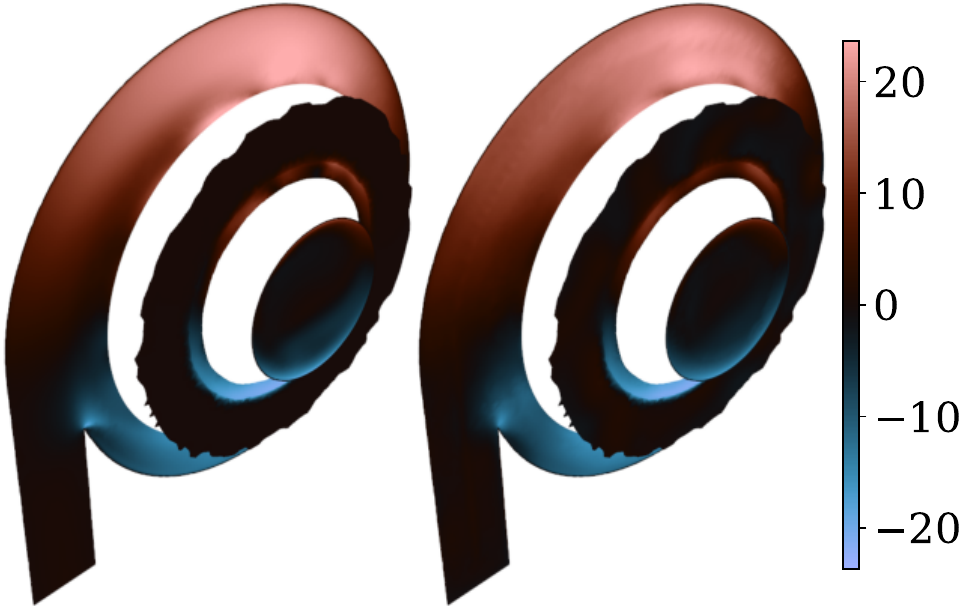}
    \caption{S1: predicted $u_x$.}
    \label{fig:pump-s1-vel-x}
\end{subfigure}\hfill
\begin{subfigure}[b]{0.2\textwidth}
    \centering
    \includegraphics[width=\linewidth]{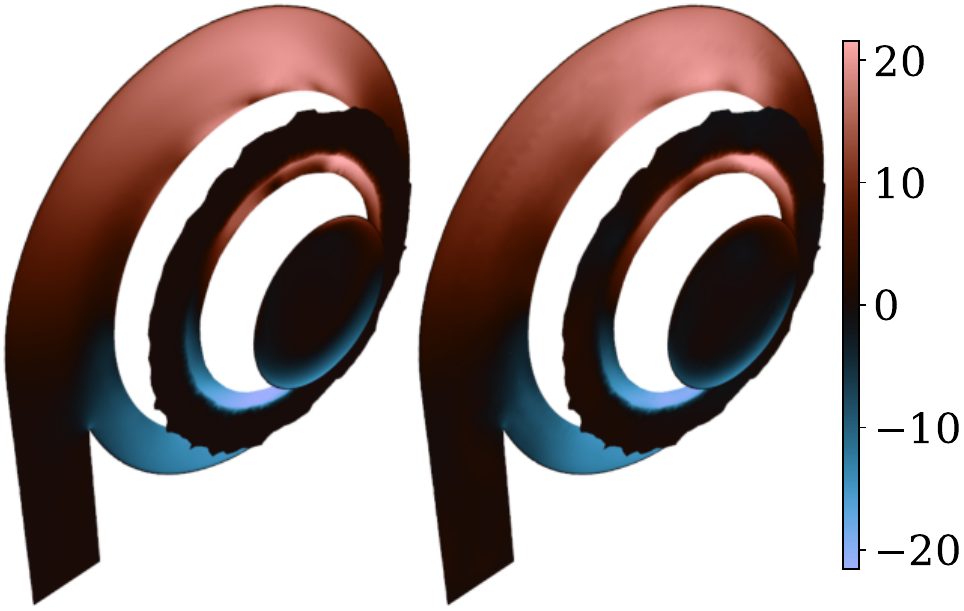}
    \caption{S2: predicted $u_x$.}
    \label{fig:pump-s2-vel-x}
\end{subfigure}\hfill
\begin{subfigure}[b]{0.2\textwidth}
    \centering
    \includegraphics[width=\linewidth]{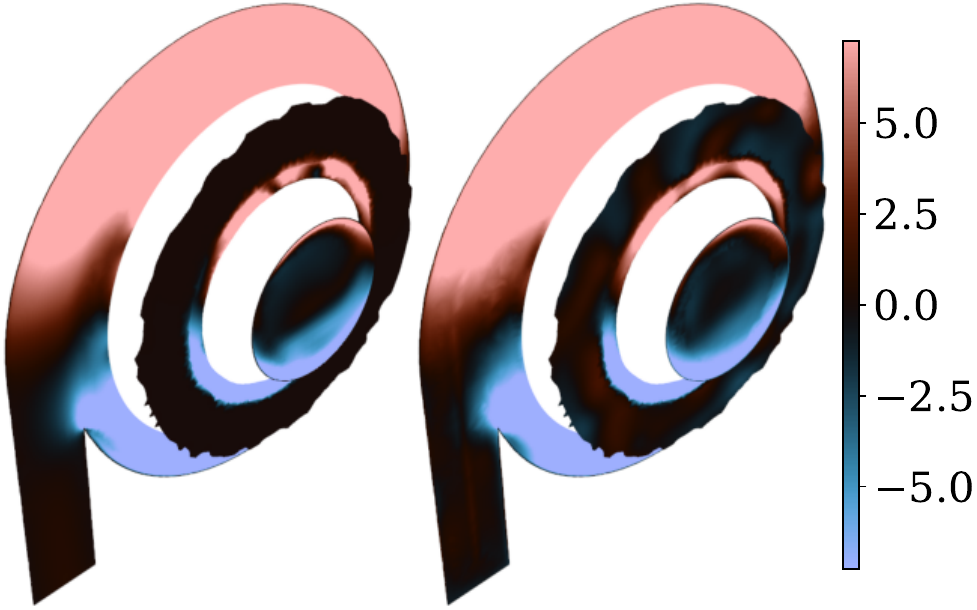}
    \caption{S1: predicted $u_y$.}
    \label{fig:pump-s1-vel-y}
\end{subfigure}\hfill
\begin{subfigure}[b]{0.2\textwidth}
    \centering
    \includegraphics[width=\linewidth]{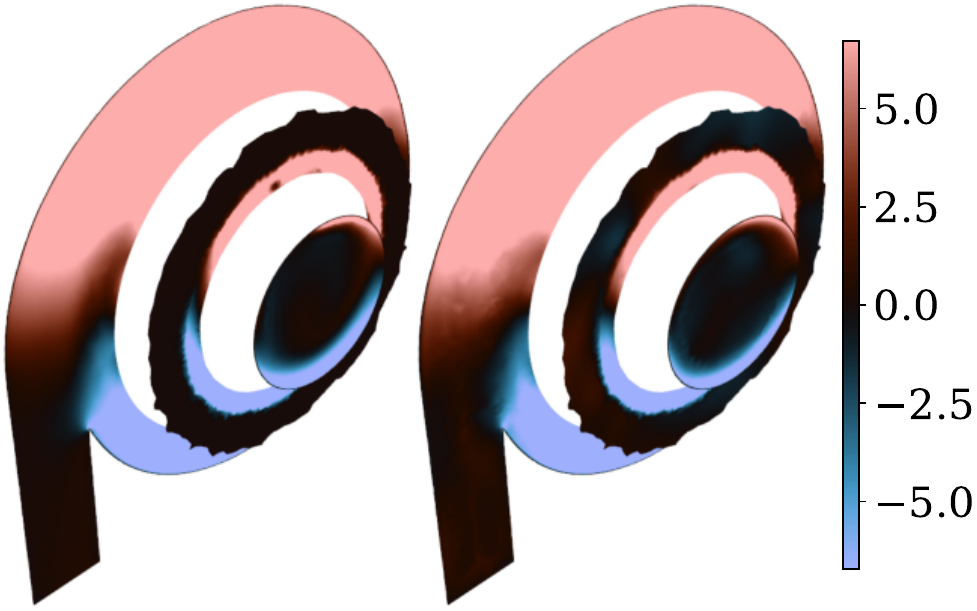}
    \caption{S2: predicted $u_y$.}
    \label{fig:pump-s2-vel-y}
\end{subfigure}
\begin{subfigure}[b]{0.2\textwidth}
    \centering
    \includegraphics[width=\linewidth]{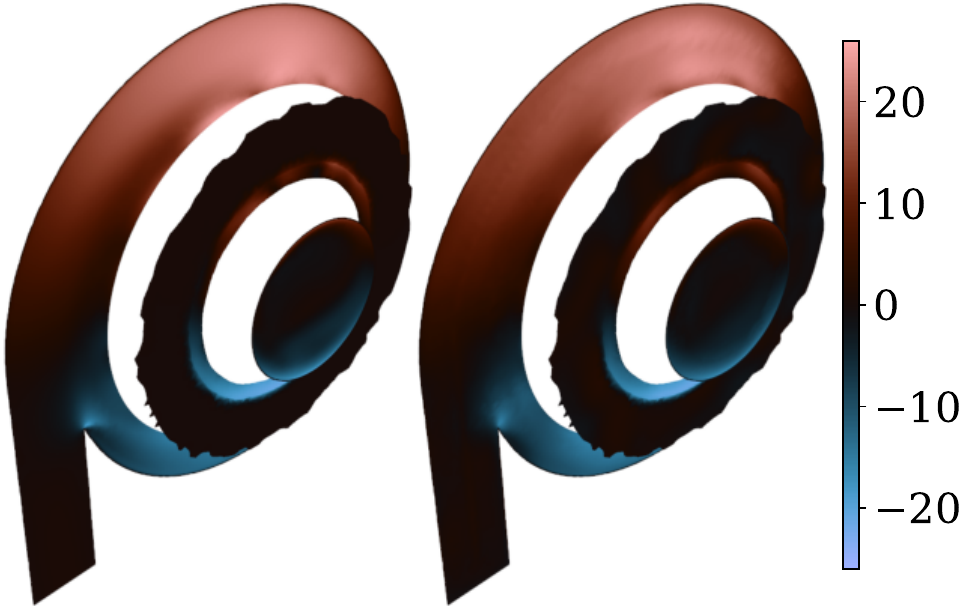}
    \caption{S1: predicted $u_z$.}
    \label{fig:pump-s1-vel-z}
\end{subfigure}\hfill
\begin{subfigure}[b]{0.2\textwidth}
    \centering
    \includegraphics[width=\linewidth]{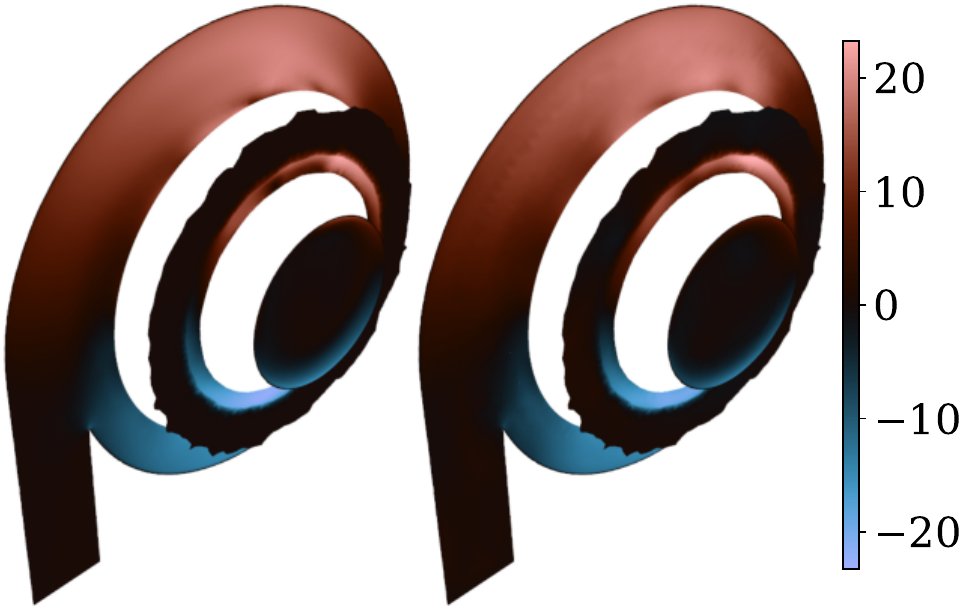}
    \caption{S2: predicted $u_z$.}
    \label{fig:pump-s2-vel-z}
\end{subfigure}\hfill
\begin{subfigure}[b]{0.2\textwidth}
    \centering
    \includegraphics[width=\linewidth]{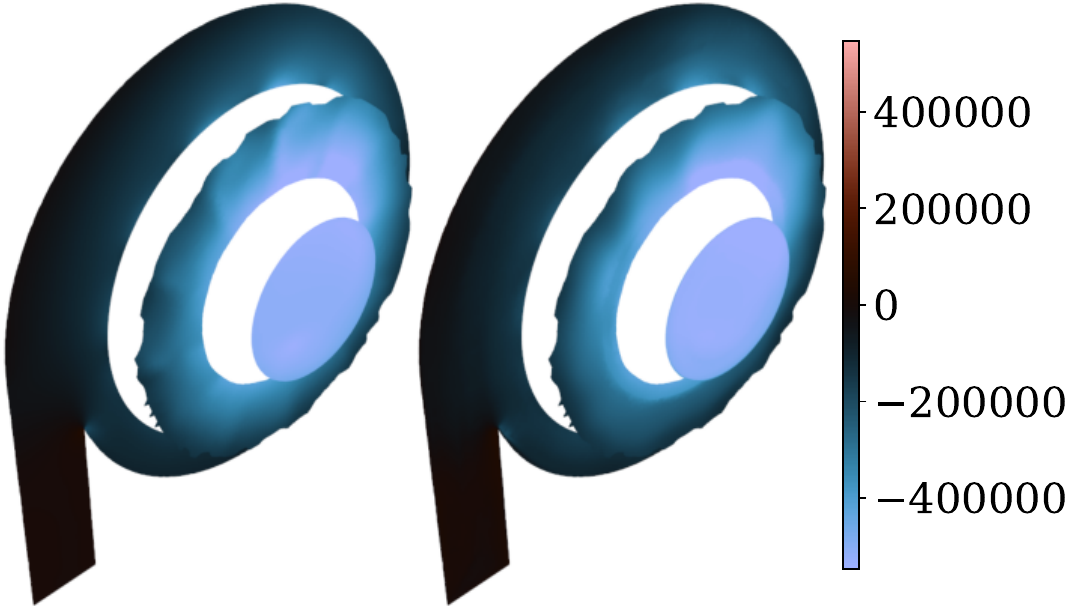}
    \caption{S1: predicted $p$.}
    \label{fig:pump-s1-pres}
\end{subfigure}\hfill
\begin{subfigure}[b]{0.2\textwidth}
    \centering
    \includegraphics[width=\linewidth]{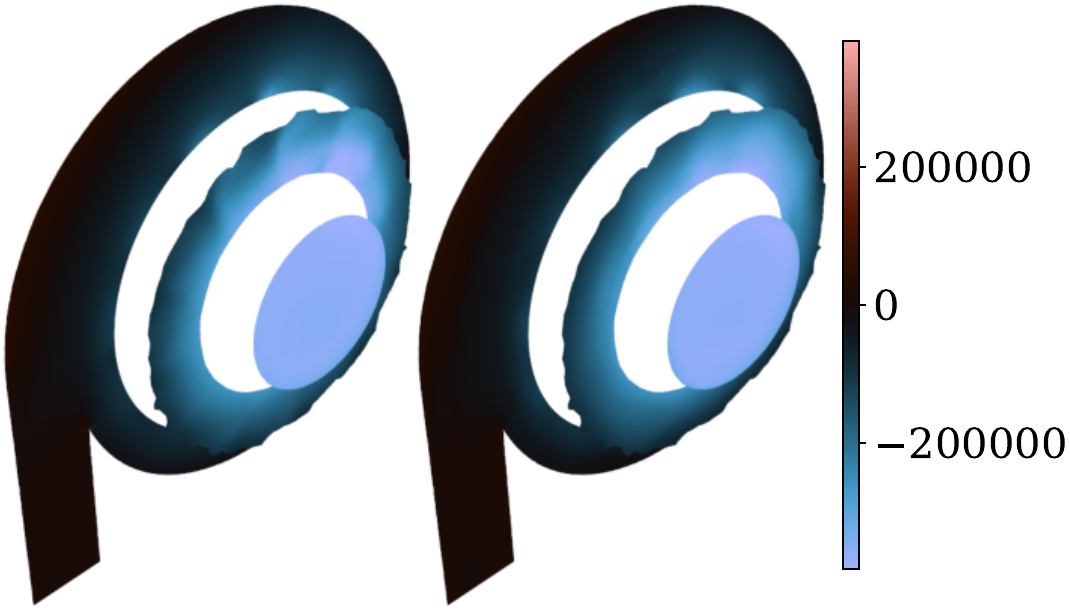}
    \caption{S2: predicted $p$.}
    \label{fig:pump-s2-pres}
\end{subfigure}
\caption{Latent basis decomposition for the 3D Centrifugal Pump
baseline model.
(a,\,b)~Ground truth and predicted fields for two test samples,
visualized at three evenly spaced cross sections.
(c,\,d)~Top-norm per-anchor decoded fields $\boldsymbol{\delta}_k$,
ranked left to right, with individual anchors specializing to distinct
regions of the geometry.
(e--l)~Full predicted fields by component ($u_x$, $u_y$, $u_z$, $p$)
for each sample.}
\label{fig:pump-flow-modes}
\end{figure}

\clearpage

\subsubsection*{With Gaussian Window Attention}
\begin{figure}[htb]
\centering
\begin{subfigure}[b]{0.4\textwidth}
    \centering
    \includegraphics[width=\linewidth]{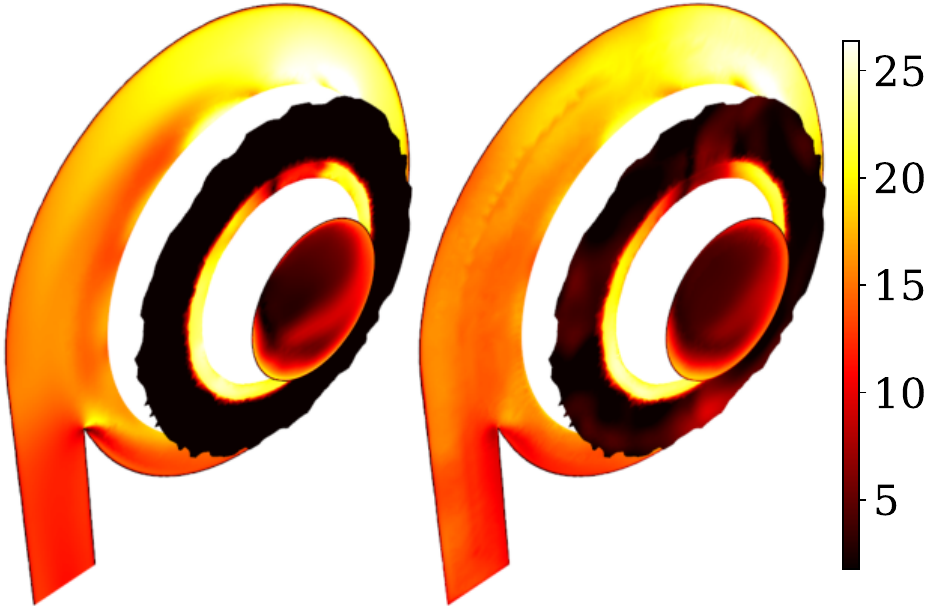}
    \caption{S1: truth vs.\ pred.}
    \label{fig:pump-gw-s1-truepred}
\end{subfigure}
\begin{subfigure}[b]{0.4\textwidth}
    \centering
    \includegraphics[width=\linewidth]{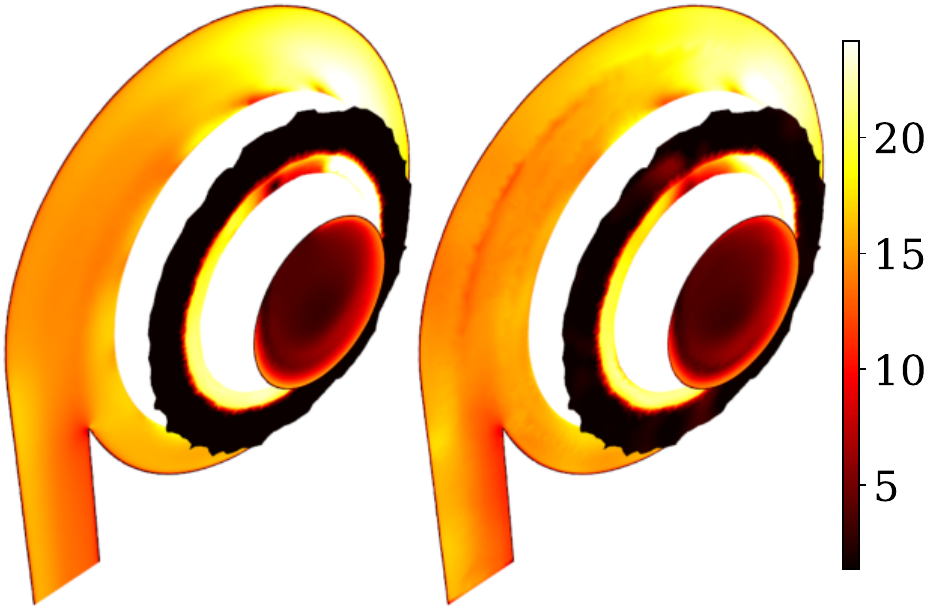}
    \caption{S2: truth vs.\ pred.}
    \label{fig:pump-gw-s2-truepred}
\end{subfigure}
\begin{subfigure}[b]{1.0\textwidth}
    \centering
    \includegraphics[width=\linewidth]{figures/appendix/pump_per_latent_decode_sample1_norm_head_basis_gw.pdf}
    \caption{S1: top-$\|\boldsymbol{\delta}_k\|$ decomposition.}
    \label{fig:pump-gw-s1-norm}
\end{subfigure}
\begin{subfigure}[b]{1.0\textwidth}
    \centering
    \includegraphics[width=\linewidth]{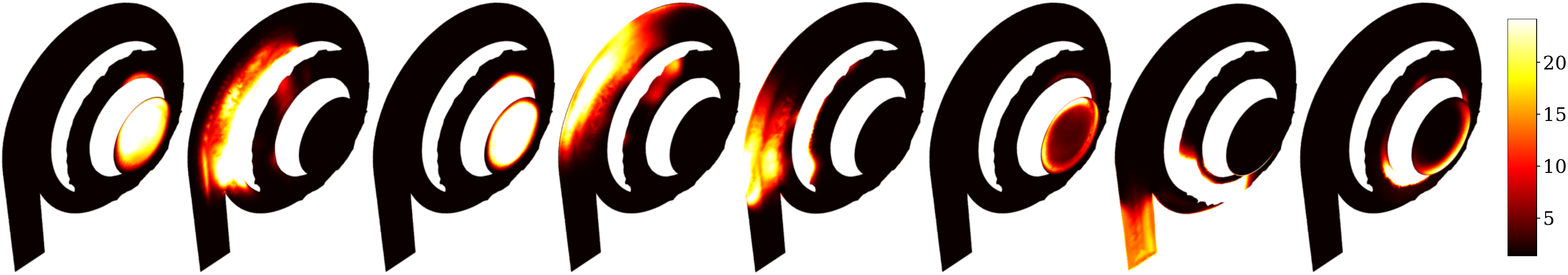}
    \caption{S2: top-$\|\boldsymbol{\delta}_k\|$ decomposition.}
    \label{fig:pump-gw-s2-norm}
\end{subfigure}
\begin{subfigure}[b]{0.2\textwidth}
    \centering
    \includegraphics[width=\linewidth]{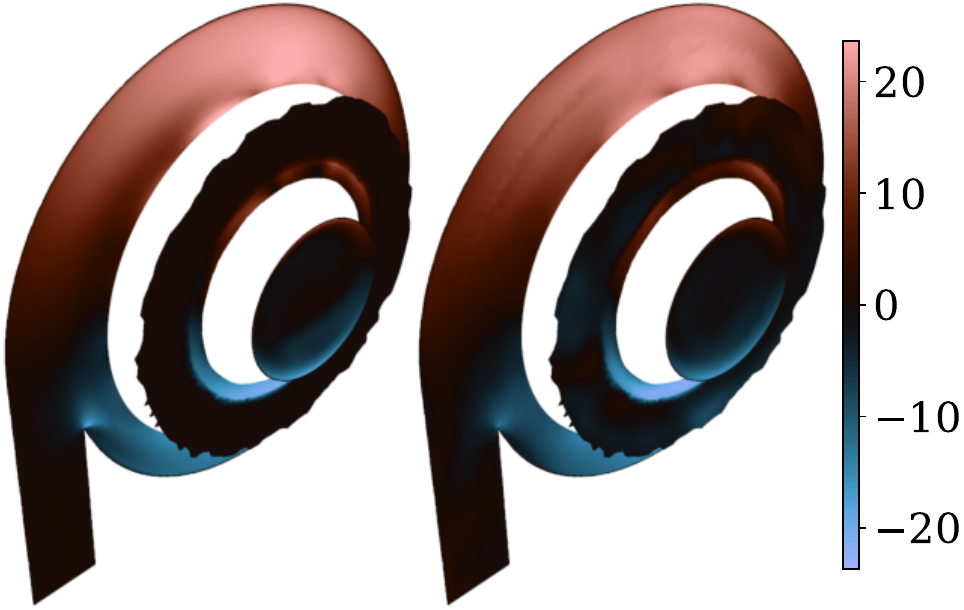}
    \caption{S1: predicted $u_x$.}
    \label{fig:pump-gw-s1-vel-x}
\end{subfigure}\hfill
\begin{subfigure}[b]{0.2\textwidth}
    \centering
    \includegraphics[width=\linewidth]{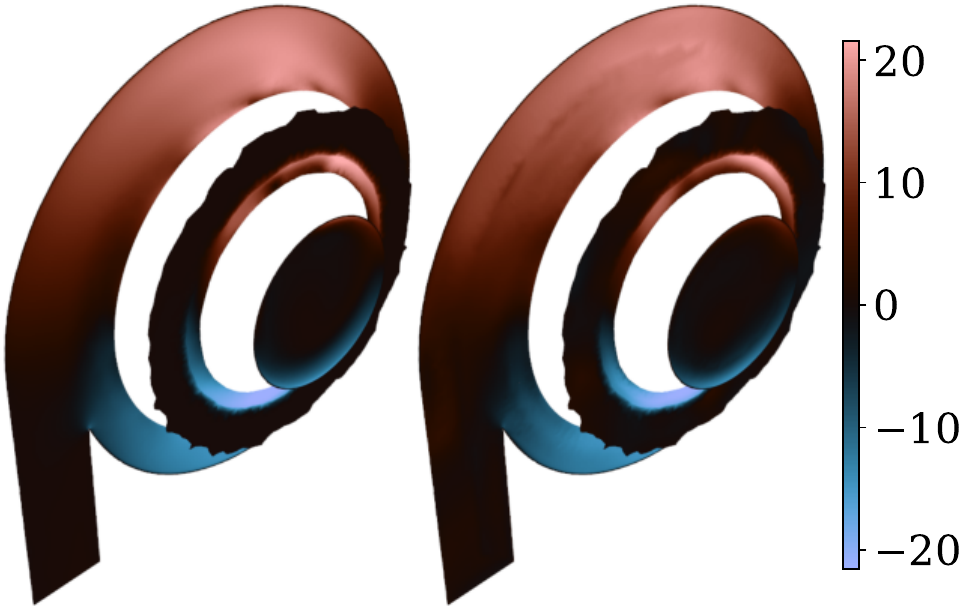}
    \caption{S2: predicted $u_x$.}
    \label{fig:pump-gw-s2-vel-x}
\end{subfigure}\hfill
\begin{subfigure}[b]{0.2\textwidth}
    \centering
    \includegraphics[width=\linewidth]{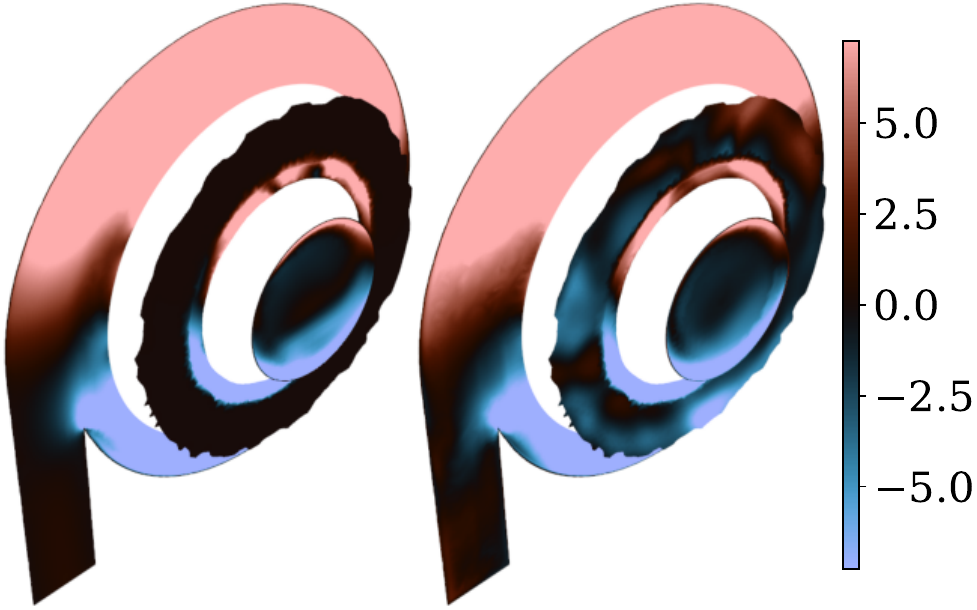}
    \caption{S1: predicted $u_y$.}
    \label{fig:pump-gw-s1-vel-y}
\end{subfigure}\hfill
\begin{subfigure}[b]{0.2\textwidth}
    \centering
    \includegraphics[width=\linewidth]{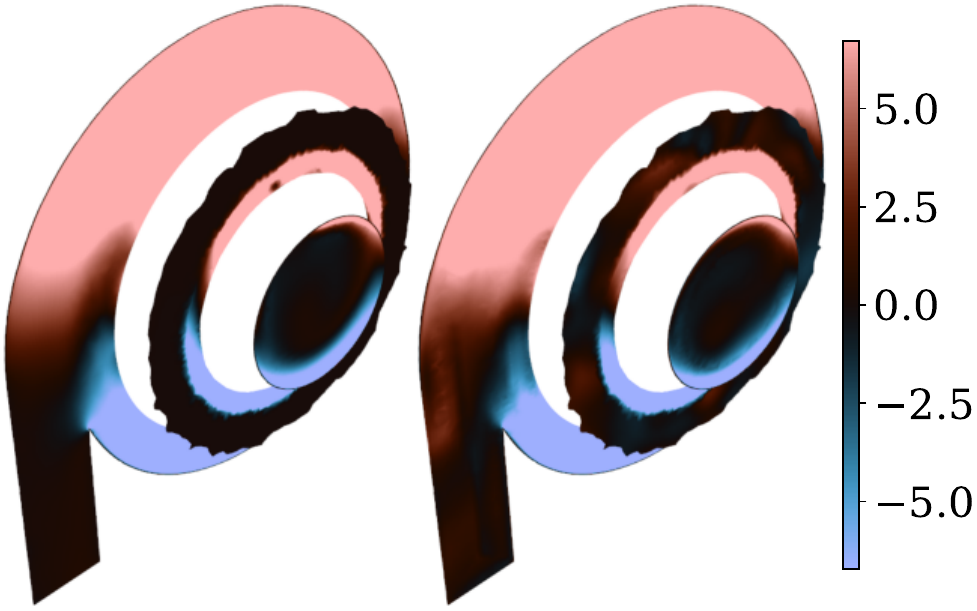}
    \caption{S2: predicted $u_y$.}
    \label{fig:pump-gw-s2-vel-y}
\end{subfigure}
\begin{subfigure}[b]{0.2\textwidth}
    \centering
    \includegraphics[width=\linewidth]{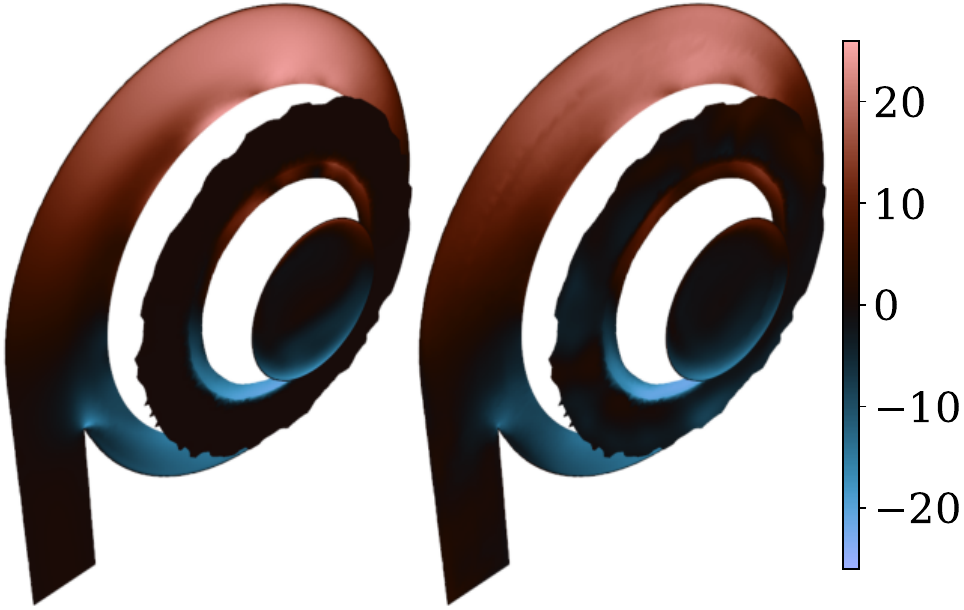}
    \caption{S1: predicted $u_z$.}
    \label{fig:pump-gw-s1-vel-z}
\end{subfigure}\hfill
\begin{subfigure}[b]{0.2\textwidth}
    \centering
    \includegraphics[width=\linewidth]{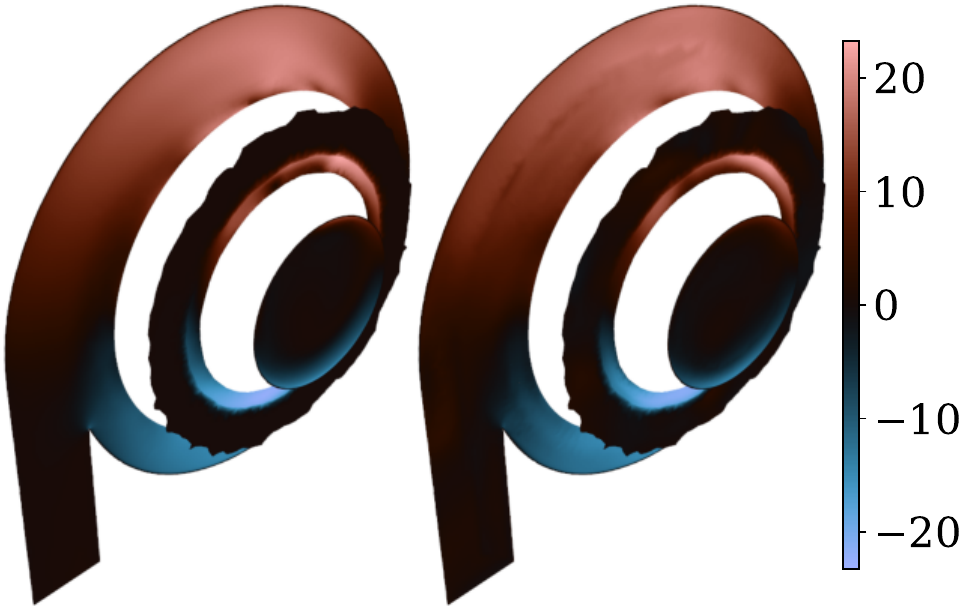}
    \caption{S2: predicted $u_z$.}
    \label{fig:pump-gw-s2-vel-z}
\end{subfigure}\hfill
\begin{subfigure}[b]{0.2\textwidth}
    \centering
    \includegraphics[width=\linewidth]{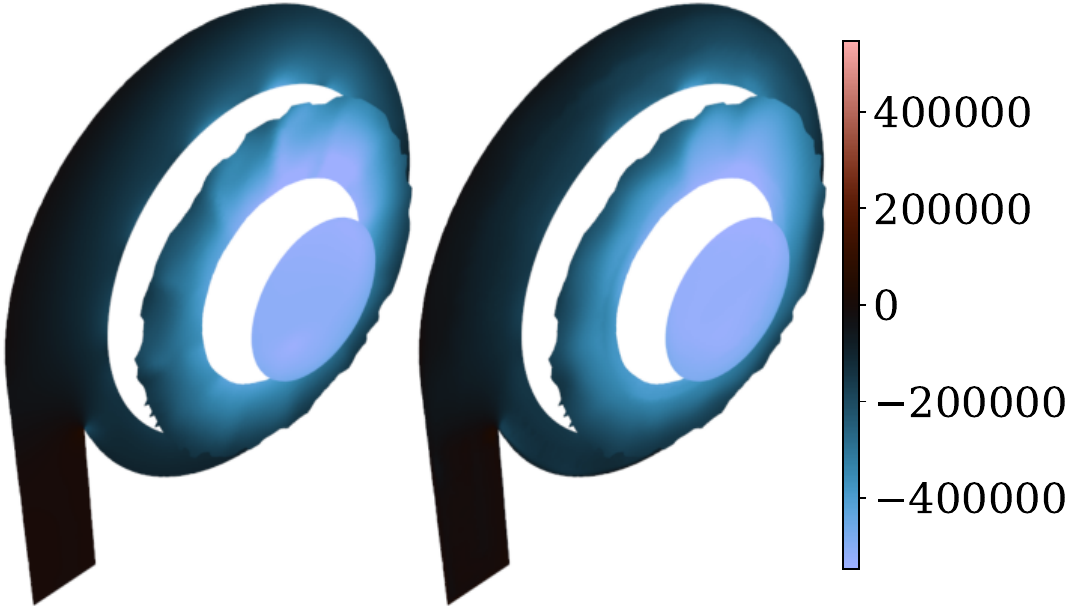}
    \caption{S1: predicted $p$.}
    \label{fig:pump-gw-s1-pres}
\end{subfigure}\hfill
\begin{subfigure}[b]{0.2\textwidth}
    \centering
    \includegraphics[width=\linewidth]{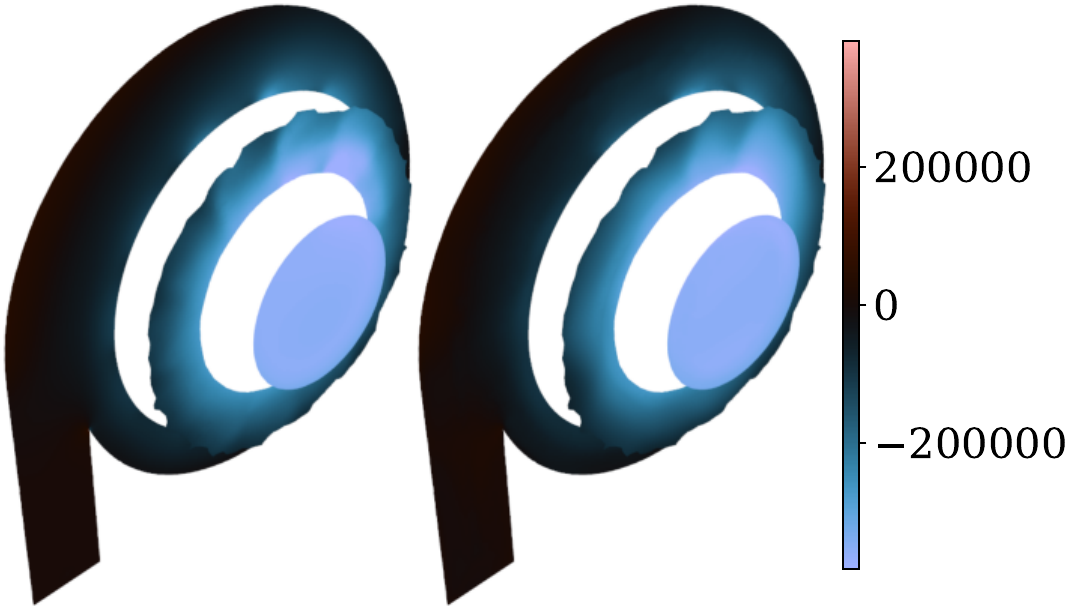}
    \caption{S2: predicted $p$.}
    \label{fig:pump-gw-s2-pres}
\end{subfigure}
\caption{Latent basis decomposition for the 3D Centrifugal Pump with
GWA modulation (Appendix~\ref{sec:gwa}).
(a,\,b)~Ground truth and predicted fields for two test samples.
(c,\,d)~Top-norm per-anchor decoded fields $\boldsymbol{\delta}_k$.
Compared with the baseline (Fig.~\ref{fig:pump-flow-modes}), GWA
produces more tightly localized per-anchor contributions.
(e--l)~Full predicted fields by component ($u_x$, $u_y$, $u_z$,
$p$).}
\label{fig:pump-flow-modes-gw}
\end{figure}

\clearpage
\subsection{3D Branched Pipe}\label{sec:3d-branched-pipe-visualizations}

\begin{figure}[htb]
\centering
\begin{subfigure}[b]{0.45\textwidth}
    \centering
    \includegraphics[width=\linewidth]{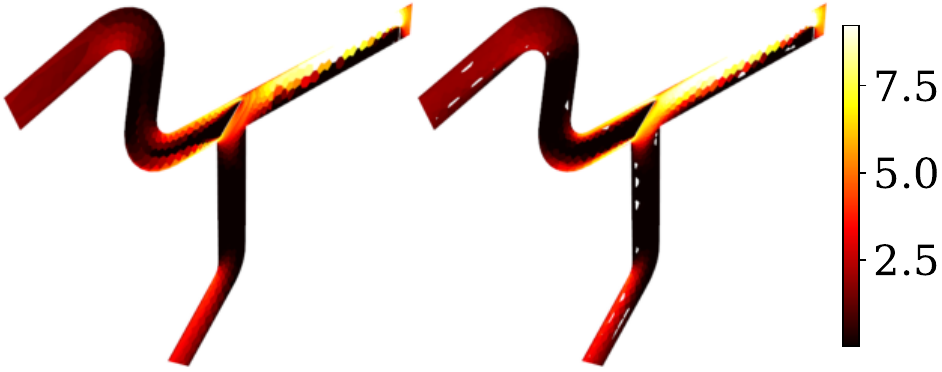}
    \caption{Sample 1: truth vs.\ pred.}
    \label{fig:pipe-s1-truepred}
\end{subfigure}\medskip
\begin{subfigure}[b]{0.45\textwidth}
    \centering
    \includegraphics[width=\linewidth]{figures/appendix/pipe_per_latent_decode_sample2_true_pred_head_basis.pdf}
    \caption{Sample 2: truth vs.\ pred.}
    \label{fig:pipe-s2-truepred}
\end{subfigure}
\begin{subfigure}[b]{0.45\textwidth}
    \centering
    \includegraphics[width=\linewidth]{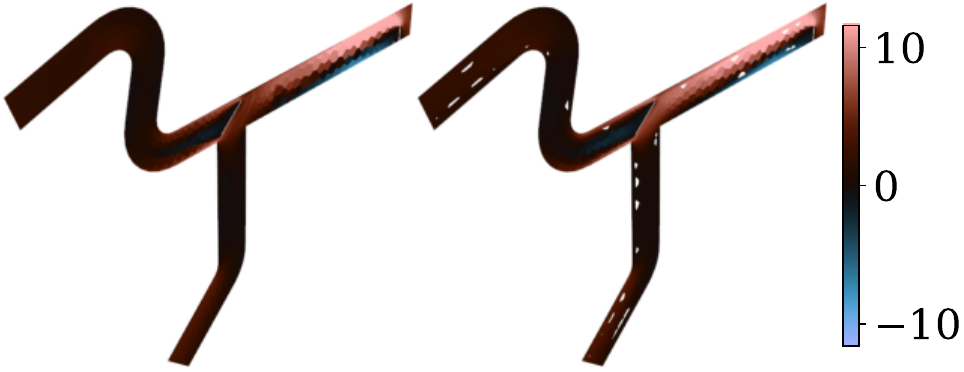}
    \caption{Sample 1: predicted $u_x$ component.}
    \label{fig:pipe-s1-vel-x}
\end{subfigure}\medskip
\begin{subfigure}[b]{0.45\textwidth}
    \centering
    \includegraphics[width=\linewidth]{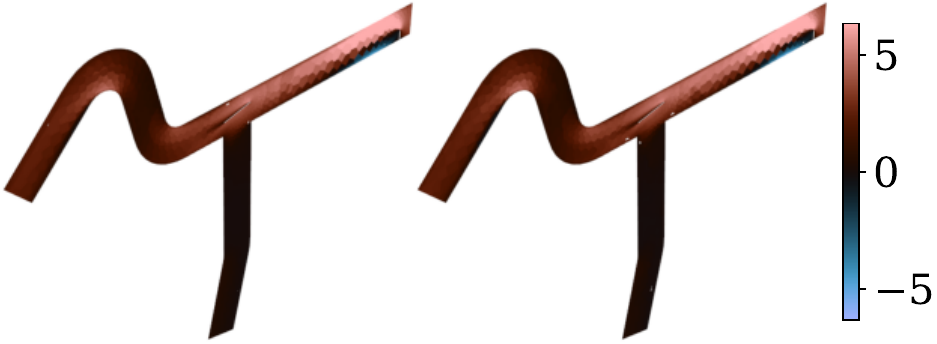}
    \caption{Sample 2: predicted $u_x$ component.}
    \label{fig:pipe-s2-vel-x}
\end{subfigure}\hfill
\begin{subfigure}[b]{0.45\textwidth}
    \centering
    \includegraphics[width=\linewidth]{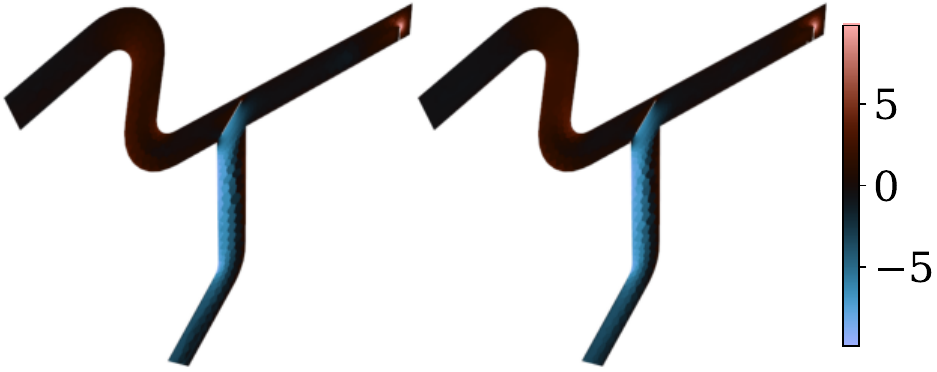}
    \caption{Sample 1: predicted $u_z$ component.}
    \label{fig:pipe-s1-vel-z}
\end{subfigure}\medskip
\begin{subfigure}[b]{0.45\textwidth}
    \centering
    \includegraphics[width=\linewidth]{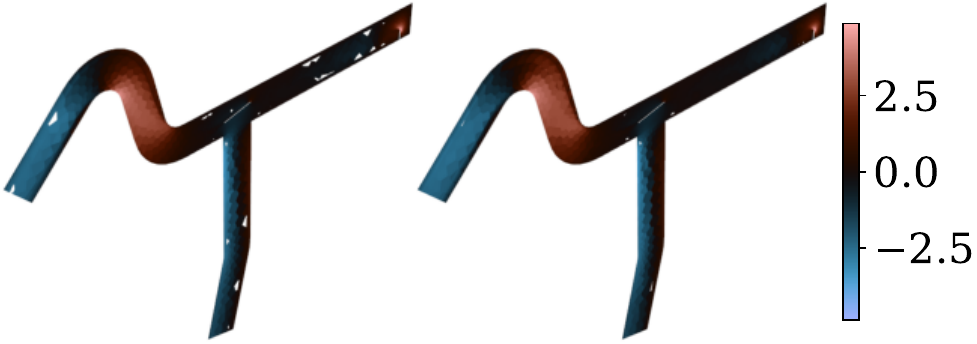}
    \caption{Sample 2: predicted $u_z$ component.}
    \label{fig:pipe-s2-vel-z}
\end{subfigure}
\hfill
\begin{subfigure}[b]{1.0\textwidth}
    \centering
    \includegraphics[width=\linewidth]{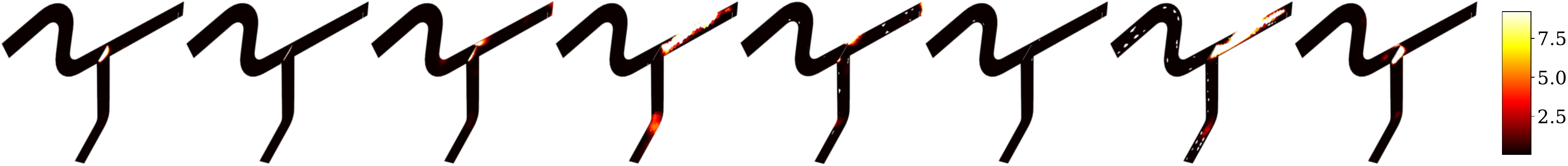}
    \caption{Sample 1: top-$\|\boldsymbol{\delta}_k\|$ decomposition.}
    \label{fig:pipe-s1-norm}
\end{subfigure}
\begin{subfigure}[b]{1.0\textwidth}
    \centering
    \includegraphics[width=\linewidth]{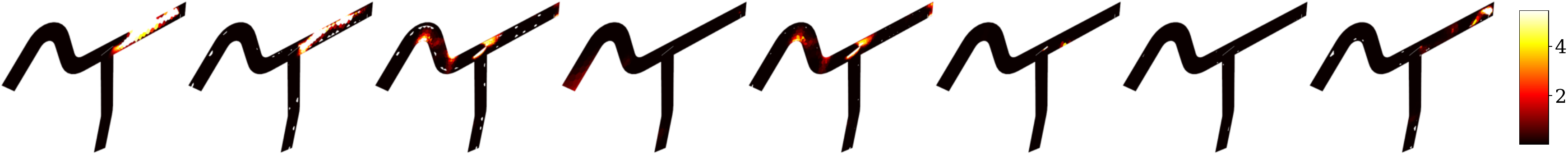}
    \caption{Sample 2: top-$\|\boldsymbol{\delta}_k\|$ decomposition.}
    \label{fig:pipe-s2-norm}
\end{subfigure}
\medskip
\caption{Latent basis decomposition for the 3D Branched Pipe Flow
baseline model.
(a,\,b)~Ground truth and predicted velocity for two test samples,
visualized at three evenly spaced cross sections.
(c--f)~Full predicted velocity field shown by component ($u_x$,
$u_z$) for each sample.
(g,\,h)~Top-norm per-anchor decoded fields $\boldsymbol{\delta}_k$,
ranked left to right; each anchor's contribution is visualized at the
cross section containing its highest-magnitude point. The decomposed
contributions are spatially compact and partition distinct flow
regions.}
\label{fig:pipe-flow-modes}
\end{figure}

\clearpage
\subsubsection*{With Gaussian Window Attention}
\begin{figure}[htb]
\centering
\begin{subfigure}[b]{0.45\textwidth}
    \centering
    \includegraphics[width=\linewidth]{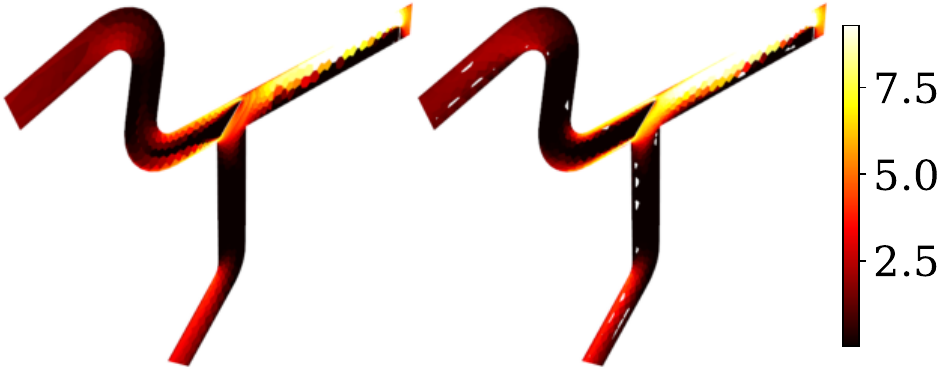}
    \caption{Sample 1: truth vs.\ pred.}
    \label{fig:pipe-gw-s1-truepred}
\end{subfigure}\medskip
\begin{subfigure}[b]{0.45\textwidth}
    \centering
    \includegraphics[width=\linewidth]{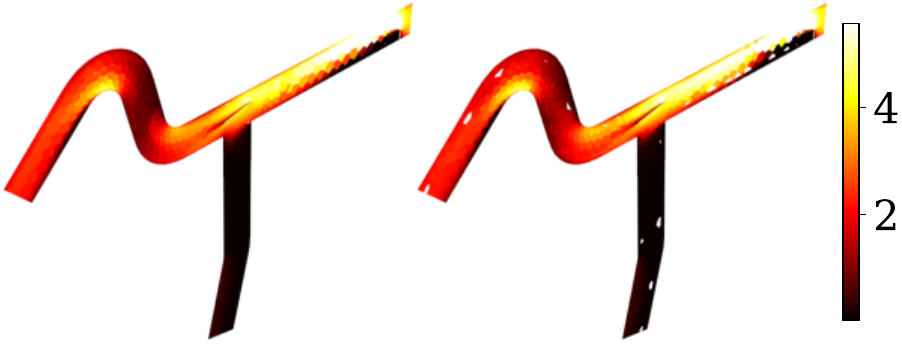}
    \caption{Sample 2: truth vs.\ pred.}
    \label{fig:pipe-gw-s2-truepred}
\end{subfigure}
\begin{subfigure}[b]{0.45\textwidth}
    \centering
    \includegraphics[width=\linewidth]{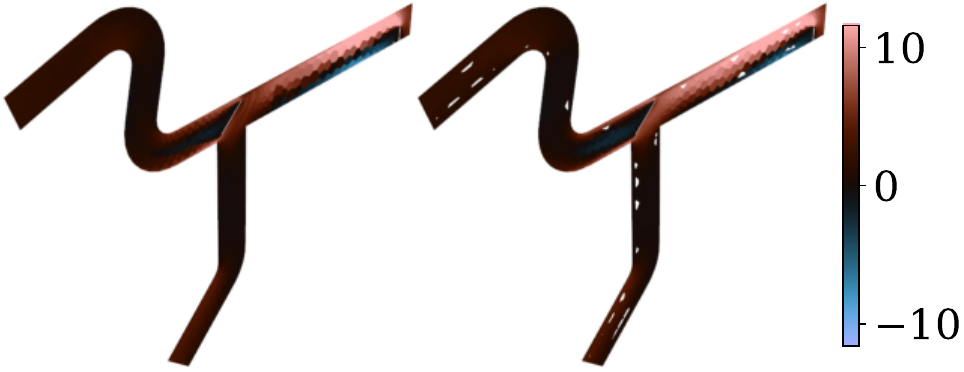}
    \caption{Sample 1: predicted $u_x$ component.}
    \label{fig:pipe-gw-s1-vel-x}
\end{subfigure}\medskip
\begin{subfigure}[b]{0.45\textwidth}
    \centering
    \includegraphics[width=\linewidth]{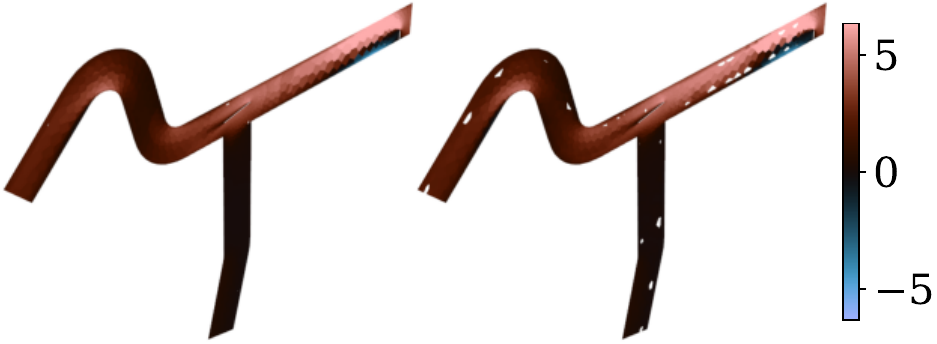}
    \caption{Sample 2: predicted $u_x$ component.}
    \label{fig:pipe-gw-s2-vel-x}
\end{subfigure}\hfill
\begin{subfigure}[b]{0.45\textwidth}
    \centering
    \includegraphics[width=\linewidth]{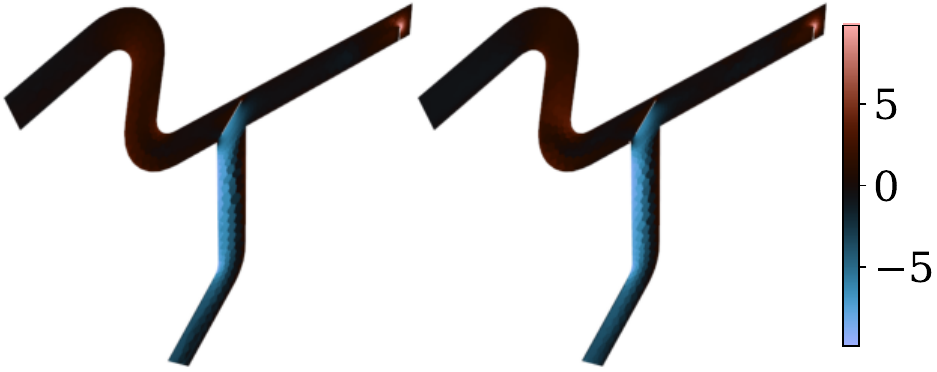}
    \caption{Sample 1: predicted $u_z$ component.}
    \label{fig:pipe-gw-s1-vel-z}
\end{subfigure}\medskip
\begin{subfigure}[b]{0.45\textwidth}
    \centering
    \includegraphics[width=\linewidth]{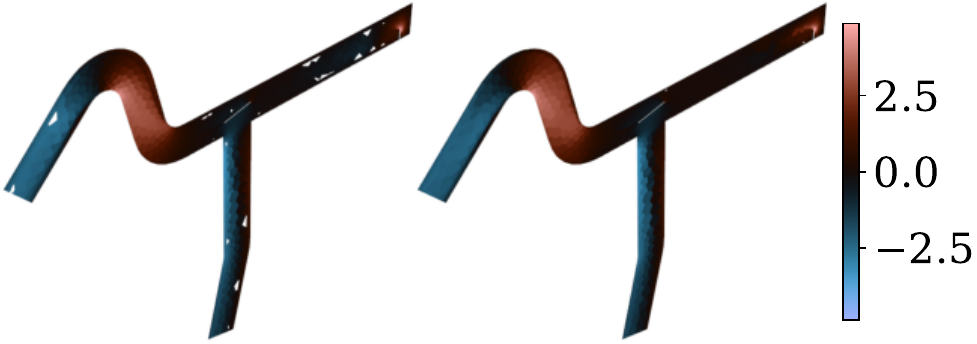}
    \caption{Sample 2: predicted $u_z$ component.}
    \label{fig:pipe-gw-s2-vel-z}
\end{subfigure}
\hfill
\begin{subfigure}[b]{1.0\textwidth}
    \centering
    \includegraphics[width=\linewidth]{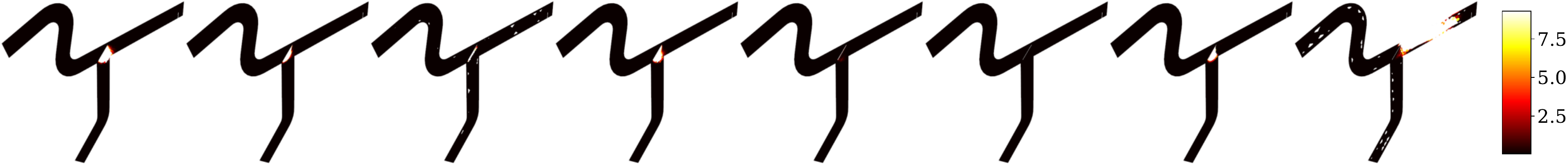}
    \caption{Sample 1: top-$\|\boldsymbol{\delta}_k\|$ decomposition.}
    \label{fig:pipe-gw-s1-norm}
\end{subfigure}
\begin{subfigure}[b]{1.0\textwidth}
    \centering
    \includegraphics[width=\linewidth]{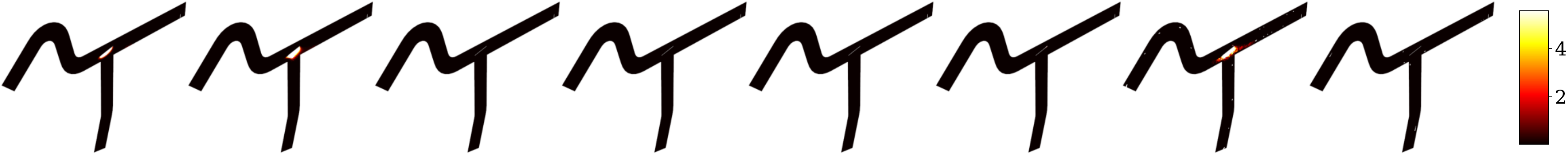}
    \caption{Sample 2: top-$\|\boldsymbol{\delta}_k\|$ decomposition.}
    \label{fig:pipe-gw-s2-norm}
\end{subfigure}
\medskip
\caption{Latent basis decomposition for the 3D Branched Pipe Flow
with GWA modulation (Appendix~\ref{sec:gwa}).
(a,\,b)~Ground truth and predicted velocity for two test samples.
(c--f)~Full predicted velocity field by component ($u_x$, $u_z$).
(g,\,h)~Top-norm per-anchor decoded fields $\boldsymbol{\delta}_k$.
Compared with the baseline (Fig.~\ref{fig:pipe-flow-modes}), GWA
produces visibly more compact spatial support for each anchor,
consistent with the distance-dependent attention bias confining each
latent to a tighter neighborhood around its anchor position.}
\label{fig:pipe-flow-modes-gw}
\end{figure}

\end{document}